%% file: main.tex
\documentclass{article} 
\usepackage{arxiv}
\setcitestyle{authoryear,open={(},close={)}}

\usepackage{fullpage}

\input{math_commands.tex}

\usepackage{url}

\usepackage{graphicx}
\usepackage{wrapfig}
\usepackage{subfig}
\usepackage{algorithm}
\usepackage[noend]{algorithmic}
\usepackage{bbm}

\usepackage{amsmath}
\usepackage{amssymb}
\usepackage{mathtools}
\usepackage{amsthm}
\usepackage{enumitem}
\usepackage{natbib}
\usepackage{xcolor}

\theoremstyle{plain}
\newtheorem{theorem}{Theorem}[section]

\theoremstyle{definition}

\theoremstyle{remark}

\title{\textbf{Bifurcated Generative Flow Networks}}

\author{
{\large Chunhui Li$^{1}$ ~ Cheng-Hao Liu$^{2}$ ~ Dianbo Liu$^{3}$ ~ Qingpeng Cai$^{4}$ ~ Ling Pan$^{1}$\thanks{Correspondence to \texttt{lingpan@ust.hk}.}} \\
 \\
{\large {$^{1}$ Hong Kong University of Science and Technology}} \\
{\large $^{2}$ Mila - Qu\'ebec AI Institute $^{3}$ National University of Singapore $^{4}$ Kuaishou Technology}
}

\begin{document}

\maketitle

\begin{abstract}
Generative Flow Networks (GFlowNets), a new family of probabilistic samplers, have recently emerged as a promising framework for learning stochastic policies that generate high-quality and diverse objects proportionally to their rewards. However, existing GFlowNets often suffer from low data efficiency due to the direct parameterization of edge flows or reliance on backward policies that may struggle to scale up to large action spaces. In this paper, we introduce Bifurcated GFlowNets (BN), a novel approach that employs a bifurcated architecture to factorize the flows into separate representations for state flows and edge-based flow allocation. This factorization enables BN to learn more efficiently from data and better handle large-scale problems while maintaining the convergence guarantee. Through extensive experiments on standard evaluation benchmarks, we demonstrate that BN significantly improves learning efficiency and effectiveness compared to strong baselines.
\end{abstract}

\section{Introduction}
Generative Flow Networks (GFlowNets)~\citep{bengio2021flow} are a new family of probabilistic samplers that learn stochastic policies aiming to sample composite objects with probability proportional to the rewards, which offer a promising approach for discovering novel, high-quality, and diverse objects in various important domains.
There have been recent advancements demonstrating the effectiveness of GFlowNets in challenging problems, e.g., molecule discovery~\citep{bengio2021flow}, biological sequence design~\citep{jain2022biological}, Bayesian structure learning~\citep{deleu2022bayesian, nishikawa2022bayesian}, combinatorial optimization~\citep{zhang2023robust,zhang2024let}, and large language models~\citep{li2023gflownets,hu2023amortizing}.

GFlowNets consider learning the policy in the flow networks (which is analogous to water flow). Its learning criterion is based on the flow consistency constraint in the flow network, which requires that the incoming flows from all parent states match the outgoing flows to all child states for each state.
To achieve this, the popular Flow Matching~\citep{bengio2021flow} objective parameterizes edge flows $F(s \rightarrow s')$ for optimization, which directly estimates the state-next state flow for each state.
However, this poses challenges for efficient learning, as accurately capturing the state-next state flow values for all possible transition pairs is difficult, and can lead to low data efficiency.

\begin{figure}[!h]
\centering
\includegraphics[width=1.0\linewidth]{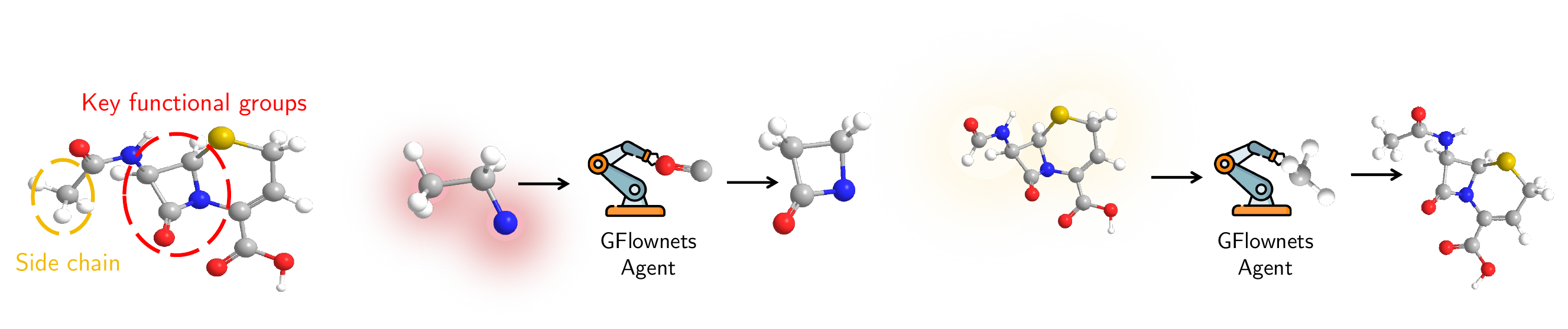}
\caption{An illustrative example in $\beta$-lactam antibiotics. Here, we highlight a key action (red) that forms the crucial $\beta$-lactam ring which is essential for antibacterial properties, versus a secondary action (yellow) that modulates the side chain to refine specific properties without drastically altering fundamental activity. }
\label{fig:intro_mol}
\end{figure}

Consider the molecule generation task~\citep{bengio2021flow} which has vast state and action spaces. 
Accurately capturing the edge flow for every possible action across all states may not always be necessary, 
if taking different actions does not affect the property a lot.
However, at certain critical states, different actions can lead to the generation of molecules with completely distinct properties,
and the impact of different actions becomes more significant. 
A representative example is $\beta$-lactam antibiotics, such as penicillins and cephalosporins~\citep{fisher2005bacterial}, shown in the left part in Figure~\ref{fig:intro_mol}. These antibiotics share a common $\beta$-lactam ring, which is crucial for their antibacterial activity by inhibiting cell wall synthesis in bacteria. 
Despite having different amide side chains, which can enhance their stability or spectrum of activity, the core $\beta$-lactam ring is the key functional group responsible for their efficacy~\citep{fisher2005bacterial}. 
This highlights that at critical junctures (e.g., the formation of the $\beta$-lactam ring), specific actions 
can significantly determine the therapeutic properties of the resulting molecule, and the GFlowNets should pay key attention to these parts (as demonstrated in the middle part in Figure~\ref{fig:intro_mol}). 
Conversely, variations in the side chains (which is a less critical state) may not drastically change the fundamental activity but can refine other properties like bacterial resistance~\citep{fisher2005bacterial}. 
Therefore, the agent does not need to pay too much attention to accurately capture the edge flow value for all possible state-next state pairs (as illustrated in the right part in Figure~\ref{fig:intro_mol}), which can simplify the learning process and improve efficiency in high-dimensional spaces.
Similarly, this example can find analogues in other fields as well, including protein design (e.g. key motif and scaffold)~\citep{wang2022scaffolding} and materials design (e.g. key composition and dopants)~\citep{shi2020advanced}.
In this paper, we aim to incorporate this key insight into the training of GFlowNets~\citep{bengio2021flow} to help improve their training efficiency.

\vspace{-.3in}
\vspace{-.15in}

Although there have been recent efforts in extending GFlowNets to the edge or trajectory level that do not explicitly directly learn edge flows~\citep{malkin2022trajectory,bengio2023gflownet,madan2023learning} since the development of the Flow Matching objective, they require modeling the backward policy $P_B$, which is hard to specify or optimize in large action space~\citep{zhang2022generative}.

To improve the learning efficiency and scalability of GFlowNets in high-dimensional problems, we propose Bifurcated GFlowNets (BN), a novel approach that factorizes the edge flows into separate representations for state flows (which estimate the flow through each state) and edge-based allocations (which determines how the flow is distributed among the outgoing edges). 
Specifically, we introduce a novel notion of edge advantage for edge-based allocations in GFlowNets, and redefine the learning objective, allowing for a more interpretable and expressive representation.
This factorization enables BN to learn more efficiently from experiences, which improves efficiency from the collected data, and scales up well to handle large state and action spaces. 

We conduct extensive experiments on standard evaluation benchmarks, including HyperGrid~\citep{bengio2021flow}, biological sequence design based on RNA sequence generation~\citep{kim2023local}, and molecule generation~\citep{bengio2021flow}. The results demonstrate that BN significantly improves learning efficiency and effectiveness compared to previous strong baselines, and can better handle large state and action spaces.

\section{Background}
\subsection{GFlowNets Preliminaries}
Given a directed acyclic graph (DAG) as $G =(\mathcal{S}, \mathcal{A})$, where $\mathcal{S}$ represents the state space and  $\mathcal{A} \subseteq \mathcal{S} \times \mathcal{S}$ represents the set of actions, we denote $s_0$ in $\mathcal{S}$ as the unique initial state with no incoming edges, and terminal states as those in  $\mathcal{X} \subseteq \mathcal{S}$ with no outgoing edges. The objective of GFlowNets is to learn a stochastic policy $P_F$ to objects $s_n$ from trajectories $\tau = (s_0 \rightarrow s_1 \rightarrow \ldots \rightarrow s_n)$ where $s_n \in \mathcal{X}$ and $(s_i \rightarrow s_{i+1}) \in \mathcal{A}$, with the sampling probability directly proportional to a reward function  $R: \mathcal{X} \rightarrow \mathbb{R}_0^+$, such that $P_{F}^{\top}(x) \propto R(x)$.  As detailed by~\citet{bengio2021flow}, in GFlowNets, the flow characterizes the transitions within the DAG. We define the trajectory flow $ F: T \rightarrow \mathbb{R}^+ $. The flow for any state $s$ is given by $ F(s) = \sum_{\tau \in T_s} F(\tau) $ where $T_s$ is the collection of all trajectories ends with state $s$. The flow for any edge $ s \rightarrow s' $ is given by $ F(s \rightarrow s') = \sum_{\tau_{s \rightarrow s'}} F(\tau) $. This setup induces a probability measure $P(\tau) = \frac{F(\tau)}{Z}$, where $Z$ represents the total flow, calculated as $Z = \sum_{\tau \in T} F(\tau)$. The forward policy $ P_F(s' | s) = \frac{F(s \rightarrow s')}{F(s)} $ and the backward policy $ P_B(s | s') = \frac{F(s \rightarrow s')}{F(s')} $ determine the transitions between states.

\subsection{Training Criteria for GFlowNets}
\noindent {\textbf{Flow Matching (FM).}} The FM~\citep{bengio2021flow} objective realizes the flow consistency constraint~\citep{bengio2021flow}, i.e., the incoming flows for a state match the outgoing flows, at the state level. 
FM directly parameterizes the edge flow function as $F_\theta(s \rightarrow s')$, where $\theta$ represents the learnable parameters. 
The optimization objective is to minimize the loss function defined in Eq.~(\ref{eq:fm_loss}) for non-terminal states, where the log-scale is used for optimization to address stability concerns~\citep{bengio2021flow}, while the objective for terminal states $x$ is to encourage the incoming flow to match the corresponding reward $R(x)$.
\begin{equation}
\mathcal{L}_{\text{FM}}(s) = (\log \sum_{s'' \in \text{Parent}(s)} F_\theta(s'' \rightarrow s) - \log \sum_{s' \in \text{Child(s)}} F_\theta(s \rightarrow s') )^2
\label{eq:fm_loss}
\end{equation}
To optimize the FM objective, trajectories are sampled from a training policy $\pi$ that ensures full support over the state space. 
This policy can be constructed as a tempered version of the forward policy $P_{F_{\theta}}$ or as a mixture of $P_{F_{\theta}}$ and a uniform policy $U$ (i.e., $\pi_{\theta} = (1 - \epsilon) P_{F_{\theta}} + \epsilon U$), which can encourage exploration and prevent premature convergence to sub-optimal solutions.
It is proven in~\citep{bengio2021flow} that if the expected loss function (Eq.~(\ref{eq:fm_loss})) reaches a global minimum and the training policy $\pi$ has full support, then FM can generate samples from the desired target distribution.

\noindent {\textbf{Detailed Balance (DB).}}
\citet{bengio2023gflownet} propose an alternative objective, DB, for achieving consistent flows in the edge level. 
The training involves learning three models: a state flow model $F_\theta(s)$, a forward policy model $P_F(\cdot | s)$, and a backward policy model $P_B(\cdot | s)$. 
The training objective of DB is to minimize the loss function defined as $\mathcal{L}_{\text{DB}}(s, s') = \left( \log(F_\theta(s) P_F(s' | s)) - \log(F_\theta(s') P_B(s | s')) \right)^2$ for non-terminal states, and a similar object is employed to encourage $F_{\theta}(x)$ to match the corresponding rewards $R(x)$ for terminal states.

\noindent {\textbf{Trajectory Balance (TB).}}
TB extends the learning objective of DB based on a telescoping calculation to directly train on full trajectories\citep{malkin2022trajectory}, whose learning objective is to minimize the loss function defined as $\mathcal{L}_{\text{TB}}(\tau) = (\log(Z_{\theta} \prod_{t=0}^{n-1} P_F(s_{t+1}|s_t)) - \log(R(x) \prod_{t=0}^{n-1} P_B(s_t|s_{t+1})) )^2$, which parameterizes the forward and backward policies and also the total flow $Z_{\theta}$.

\noindent {\textbf{Sub-Trajectory Balance (SubTB).}}
\citet{madan2023learning} propose the SubTB learning objective to mitigate the large variance problem of TB by considering the flow consistency constraint in the sub-trajectory level.
Specifically, it considers all possible $O(n^2)$ sub-trajectories $\tau_{i:j} =  \{s_i, \cdots, s_j\}$, and the optimization objective is defined as in Eq.~(\ref{eq:subtb_loss}), where $w_{ij}=\frac{\lambda^{j-i}}{\sum_{0 \leq i < j \leq n} \lambda^{j-i}}$ 
denotes the weight for $\tau_{i:j}$, and $\lambda$ represents the weighting hyperparameter.
\begin{equation}
\mathcal{L}_{\text{SubTB}}(\tau) = \sum_{\tau_{i:j} \in \tau} w_{ij} \left( \log F(s_i) \prod_{t=i}^{j-1} P_F(s_{t+1}|s_t) - \log F(s_j) \prod_{t=i}^{j-1} P_B(s_t|s_{t+1}) \right)^2
\label{eq:subtb_loss}
\end{equation}

\section{Related Work}
Generative Flow Networks (GFlowNets)~\citep{bengio2021flow} have been recently applied across various challenging domains such as molecule discovery~\citep{bengio2021flow}, biological sequence design~\citep{jain2022biological}, Bayesian structure learning~\citep{deleu2022bayesian, nishikawa2022bayesian}, combinatorial optimization~\citep{zhang2023robust,zhang2024let}, and large language models~\citep{li2023gflownets,hu2023amortizing}. These applications highlight the practical significance of GFlowNets, prompting both theoretical exploration and methodological enhancements to optimize their functionality. Several studies have provided deep theoretical insights into GFlowNets~\citep{bengio2023gflownet, malkin2022gflownets, zimmermann2022variational, zhang2022unifying, lahlou2023theory}, laying a foundation for further innovations. Following the initial Flow Matching objective by ~\citet{bengio2021flow}, there have been recent works for considering other training objectives, including the Detailed Balance~\citep{bengio2023gflownet}, Trajectory Balance~\citep{malkin2022trajectory}, and Sub-Trajectory Balance~\citep{madan2023learning}. Additionally, \cite{zhang2022generative} propose a joint training of GFlowNets with energy or reward functions, while \cite{pan2023better} develop forward-looking objectives to guide future improvements. Meanwhile, several innovative adaptations of GFlowNets have been introduced to address specific challenges~\citep{pan2022generative, pan2023stochastic, zhang2023distributional, lau2024qgfn}. \cite{pan2022generative} incorporated intrinsic exploration rewards to facilitate learning in environments with sparse rewards. \cite{pan2023stochastic} and \cite{zhang2023distributional} have tailored GFlowNets to environments with stochastic transitions and reward dynamics. 
Furthermore, \cite{pan2024pre} focused on the pre-training of GFlowNets, which proposed a reward-free, self-supervised, outcome-conditioned approach, enhancing GFlowNets' adaptability to new reward functions in downstream tasks.
In deep reinforcement learning, \cite{wang2016dueling} introduces the advantage function with Dueling DQN, which differentiates the benefit of specific actions from the general state value. This function splits the Q-value into components that separately assess the overall state and the relative benefit of each action, enhancing policy learning by focusing on the unique contribution of actions. The implementation of the advantage function has significantly boosted the efficiency and robustness of reinforcement learning algorithms.

\section{Proposed Method}
In this section, we first present a motivating example to illustrate the problem of inefficient data utilization in a popular GFlowNets approach~\citep{bengio2021flow}. Then we present our method, which is a novel approach that effectively factorizes the representations of edge flows into state flow model and edge-based allocations, allowing for more efficient learning and can better handle complex problems with large-scale action spaces.

\subsection{Motivation}
As discussed above, in certain molecular states, the choice of which atom or chemical group to add may have minimal impact on the properties of the final molecule. 
In these scenarios, the edge flows need not focus extensively on learning the relative merits of each action. 
Instead, it can concentrate on certain states where the choice of action is crucial, such as when adding a specific group at a particular position could produce toxic byproducts, or when the closure of a ring structure significantly influences the molecule's physicochemical properties.
The direct parameterization of edge flows as in Flow Matching (FM)~\citep{bengio2021flow} can therefore lead to decreased data efficiency, which we illustrate through a simple didactic example below.

\begin{wrapfigure}{r}{0.35\textwidth} \vspace{-.15in}
\centering
\includegraphics[width=1.\linewidth]{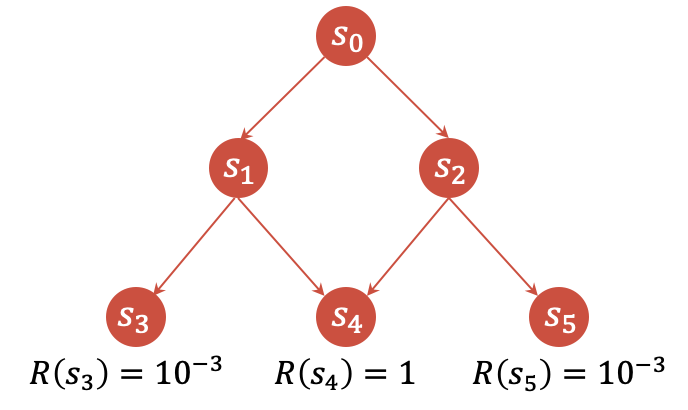}
\vspace{-.28in}
\caption{A simple scenario demonstrating the data inefficiency problem.}
\label{fig:simple_dag}
\vspace{-.5cm}
\end{wrapfigure}
Consider a simple directed acyclic graph (DAG) as shown in Figure~\ref{fig:simple_dag}, which illustrates a hard exploration problem with sparse rewards, as only $s_4$ has a reward of $1$ while other terminal states have very small terminal rewards.
Suppose that the FM agent has sampled a trajectory $\tau=\{s_0, s_2, s_5\}$ leading to $s_5$ with a reward of $R(s_5)=10^{-3}$.
According to the learning criterion of FM, the update for $\tau$ only directly affects the edge flow $F(s_2 \to s_5) = R(s_5) = 10^{-3}$. 
However, it does not effectively propagate meaningful information to other transitions ($s_2 \to s_4$), leaving them underexplored. 
In contrast, with an effective decomposition of the edge flow into separate representations for state flows and edge-based allocations (to be introduced in Section~\ref{sec:bn}), the update for a transition will update the value of the corresponding state flows, e.g., $F(s_2)$, in a meaningful way based on the reward.
This update informs all outgoing edges from $s_2$, thereby enhancing exploration by providing valuable information for potential trajectories like $s_0 \to s_2 \to s_4$.
Therefore, it can address FM's data inefficiency problem by ensuring that the training signal is more broadly disseminated throughout the underlying DAG, facilitating better exploration and discovery of modes in the reward function.
It is also worth noting that although there have been recent methods~\citep{bengio2023gflownet,malkin2022trajectory,madan2023learning} that do not explicitly model edge flows, they rely on a backward policy $P_B(s'|s)$ that can be hard to learn or specify in large scale problems with huge action spaces~\citep{zhang2022generative}, which will be discussed in Section~\ref{sec:bn}.

In the next section, we will introduce our method for achieving efficient data utilization, which enhances learning efficiency and effectively scales up to larger action space problems.

\subsection{{\underline{B}}ifurcated Generative Flow {\underline{N}}etworks (BN)} \label{sec:bn}
As discussed in the previous sections, it can pose a significant challenge for the agent to accurately capture the state-next state flows for every possible state-next state pair and for efficiently learn from experiences.
We aim to develop a novel concept in GFlowNets that naturally separates the edge flow function into a state-only dependent state flow and an edge-related component responsible for assigning importance to different actions within a single state. 
In reinforcement learning, there is a well-defined advantage function~\citep{sutton1999reinforcement}, where the $Q$-value for a state-action pair $Q(s,a)$ can be decomposed into the summation of the state-value $V(s)$ and advantage function $A(s,a)$, i.e., $Q(s,a) = V(s) + A(s,a)$.
The advantage function-based $Q$-value decomposition~\citep{wang2016dueling} can greatly improve sample efficiency by more efficient information propagation.

However, extending the notion of the advantage function to GFlowNets is non-trivial. The state flow for a given state $s'$ is defined as the summation of incoming edge flows from its parent states $s$, i.e., $F(s) = \sum_{s'' \in \text{Parent}(s)} F(s'' \to s)$, and the naive definition of the advantage function can lead to negative advantage flows, which is impractical and inconsistent with the non-negative flow constraints in flow network systems.
To address this challenge, we carefully analyze the flow definition in GFlowNets and introduce a novel notion of edge-based allocation, defined as $F(s \to s') = F(s) A(s'|s)$, where $F(s)$ represents the state flow and $A(s'|s)$ represents the edge-based allocation (as a probability distribution). This decomposition allows for a more interpretable and expressive representation of the flow in GFlowNets.

Based on this notion of edge-based allocations, we derive the learning objective of {\underline{B}}ifurcated Generative Flow {\underline{N}}etworks (BN) as in Eq.~(\ref{eq:bn_obj}) by extending Flow Matching to allow for efficient decomposition of the edge flow.
\begin{equation}
\sum_{s} F(s) A(s'|s) = F(s') = \sum_{s''} F(s') A(s''|s')
\label{eq:bn_obj}
\end{equation}
It is also worth noting that the right-hand-side can be further simplified as $\sum_{s''}A(s''|s')=1$, and we obtain the final loss function for learning BN as in Eq.~(\ref{eq:bn_loss}) for non-terminal states, and a similar objective encouraging the inflow at terminal states $x$ equal the reward $R(x)$.
\begin{equation}
\mathcal{L}_{\text{BN}}(s') = \left( \log \sum_{s \to s' \in \mathcal{A}} F(s) A(s'|s) - \log F(s') \right)^2
\label{eq:bn_loss}
\end{equation}

\paragraph{Theoretical justification.}
In Theorem~\ref{thm}, we provide a theoretical justification for the BN learning objective, demonstrating its convergence and correctness. The detailed proof can be found in Appendix~\ref{app:proof}.
\begin{theorem}
If $\mathcal{L}_{\text{BN}}(s')=0$ for all states, then the edge advantage policy $A(s'|s)$ samples proportionally to the reward function.
\label{thm}
\end{theorem}

\begin{figure}[!h]
    \centering
    \begin{minipage}{0.45\textwidth}
        \centering
        \includegraphics[width=\textwidth]{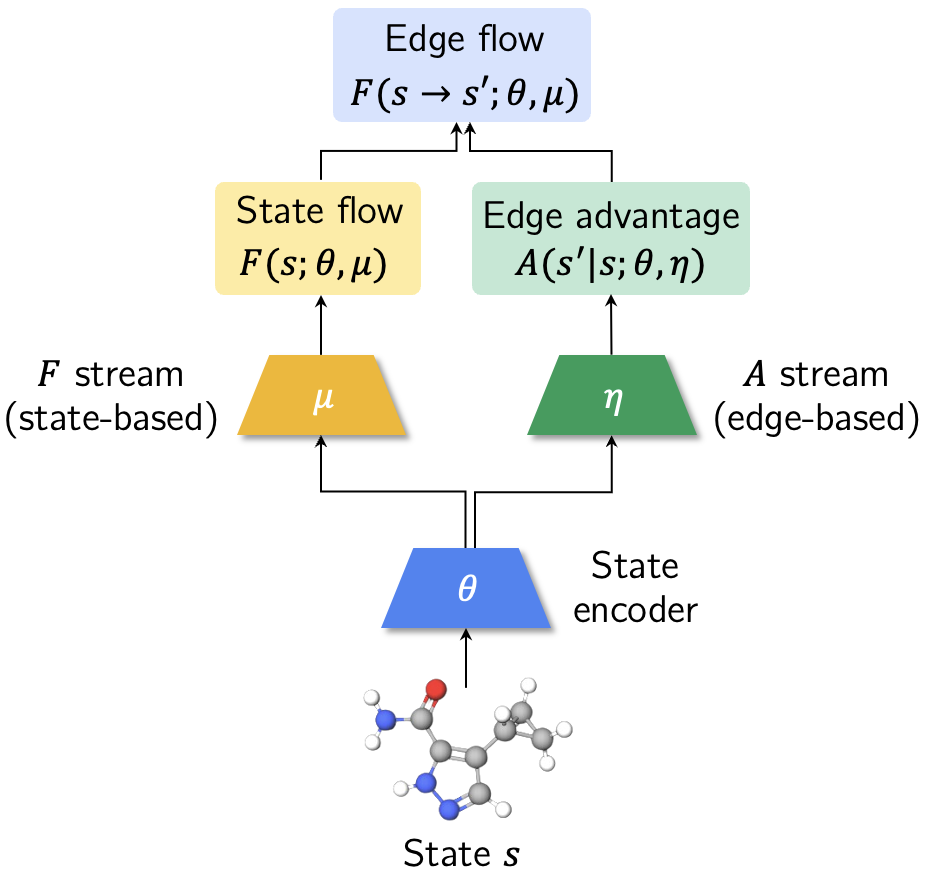}
        \caption{The network structure of BN.}
        \label{fig:bn_net}
    \end{minipage}
    \begin{minipage}{0.45\textwidth}
        \centering
        \subfloat[Small.]{\includegraphics[width=.58\linewidth]{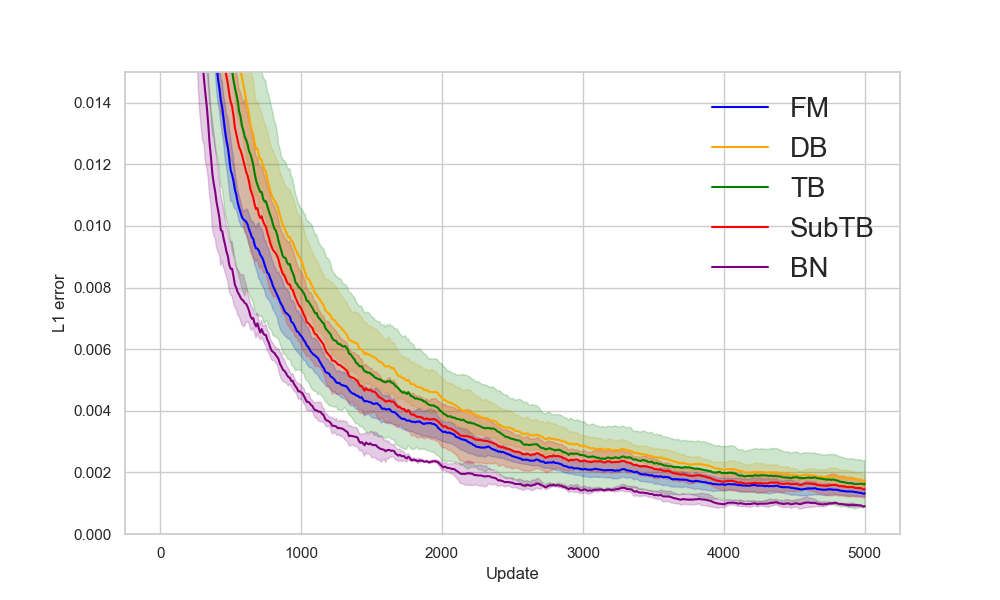}} \\
        \vspace{.07in}
        \subfloat[Large.]{\includegraphics[width=.58\linewidth]{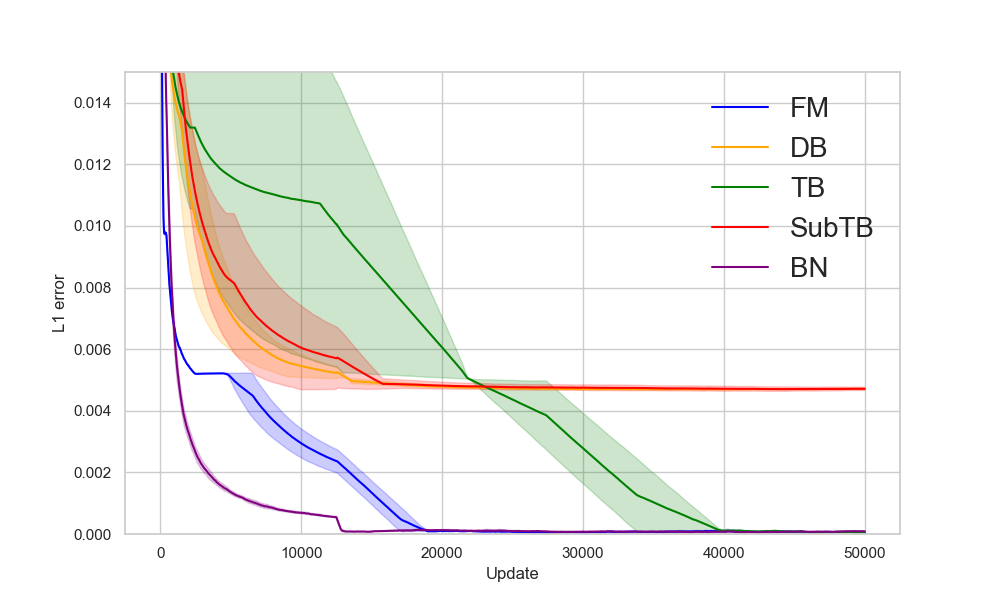}}
        \caption{Results in the didactic task.}
        \label{fig:didactic_res}
    \end{minipage}
\end{figure}

The idea can indeed be implemented in a simple yet effective manner, where Figure~\ref{fig:bn_net} illustrates the learning structure of our proposed Bifurcated GFlowNets. It consists of a carefully designed bifurcated architecture with a shared state encoder parameterized by $\theta$,
which decomposes the edge flow $F(s \to s')$ into a state flow network stream $F(s)$ parameterized by $\mu$ and an edge advantage network stream $A(s'|s)$ parameterized by $\eta$. 

\paragraph{Empirical validation.}
\label{sec:empirical validation}
We design a didactic environment to validate the effectiveness of BN, which is a tabular DAG with controllable state and action spaces and more complicated transition dynamics than Figure~\ref{fig:simple_dag}.
We consider two environment specifications, small and large, with varying state and action space sizes, where the full description of the environment can be found in Appendix~\ref{app:setup_toyDAG}.
We evaluate the empirical $L_1$ error~\citep{bengio2021flow} of our method and compare it with previous GFlowNets objectives including FM~\citep{bengio2021flow}, DB~\citep{bengio2023gflownet}, TB~\citep{malkin2022gflownets}, and SubTB~\citep{madan2023learning}.
Figure~\ref{fig:didactic_res} shows that BN learns much faster than FM, with its well-designed learning structure, where the performance gap becomes larger as the problem scale increases.
It is worth noting that as the problem size grows, all methods that rely on learning the backward policy (including DB, TB, and SubTB), face challenges in scaling effectively, which highlights the difficulty of specifying or learning the backward policy in large-scale problems~\citep{zhang2022generative}.

\paragraph{Discussion.}
Our Bifurcated Generative Flow Networks (BN) introduce a novel decomposition of the edge flow in GFlowNets~\citep{bengio2021flow}, inspired by the dueling deep Q-network (DQN)~\citep{wang2016dueling} architecture. 
However, it is non-trivial to directly extend such decomposition to GFlowNets, as it can lead to impractical negative edge-based flows.
Through an in-depth analysis of the flow function in GFlowNets, we propose an innovative edge flow decomposition that effectively separates the representations into state-related and edge-related components, which significantly enhances data efficiency.
Our method is also related to the actor-critic framework~\citep{sutton1999policy}, featuring a state-dependent component that estimates the flow value for each state and a policy-like component for edge-based allocations. In contrast, FM can be considered as a pure value-based approach, while methods like DB, TB, and SubTB involve learning a state flow function, a forward policy, and an additional backward policy. BN offers a new perspective by circumventing the complexities of the introduction of the backward policy, which can be challenging to specify or learn in high-dimensional problems.
BN is also related to expected Sarsa~\citep{van2009theoretical} which uses the expected value to estimate the target Q-value and can therefore reduce variance~\citep{van2009theoretical}. BN considers a weighted combination determined by the edge-based allocations for all the parent states for state $s'$. This approach provides a more stable learning target, aligning with the principles of expected Sarsa and contributing to improved learning stability.
In summary, our BN method offers an innovative approach that enhances data efficiency, avoids the need for learning or specifying a backward policy, and provides a new perspective compared to existing methods.

\section{Experiments}\label{sec:experiments}
In this section, we conduct extensive experiments to investigate our approach by comparing it with strong baselines to investigate its efficiency and effectiveness in standard evaluation benchmarks including HyperGrid, RNA sequence generation, and molecule generation.

\subsection{HyperGrid}
\subsubsection{Experimental Setup}
\begin{wrapfigure}{r}{0.2\textwidth} \vspace{-.15in}
\centering
\includegraphics[width=1.\linewidth]{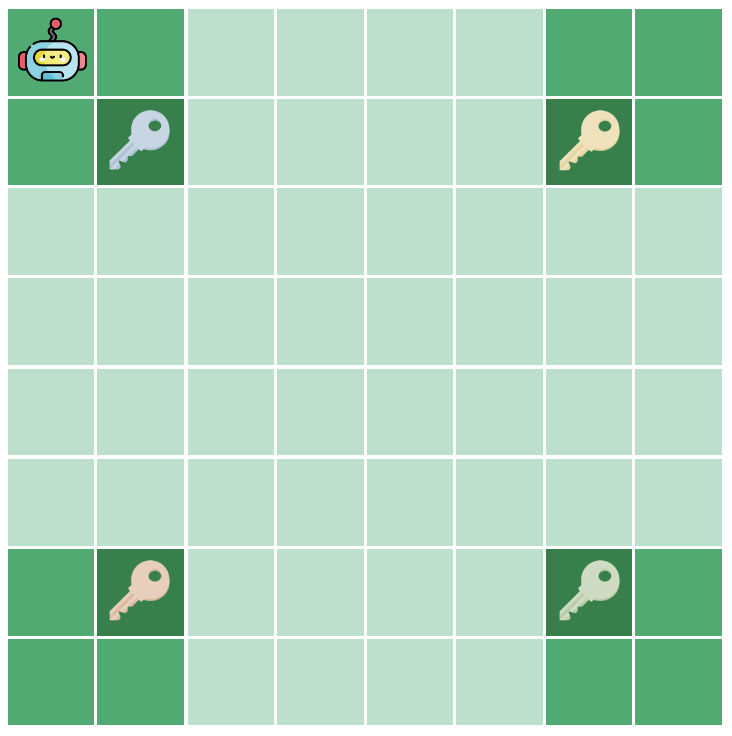}
\vspace{-.2in}
\caption{The HyperGrid task.}
\label{fig:grid_example}
\vspace{-1cm}
\end{wrapfigure}

We first evaluate our method on a toy HyperGrid task to validate its effectiveness as in~\citep{bengio2021flow}. The agent navigates in the $d$-dimensional grid-based world by selecting actions at each timestep. These actions allow the agent to move along one coordinate axis or terminate the episode when it chooses the stop action, which ensures that the underlying Markov decision process (MDP) forms a directed acyclic graph.
The agent receives a reward determined by the function $R(x)$ as defined in~\citep{bengio2021flow} upon reaching a terminal state $x$ at the end of the trajectory.
Specifically, the reward function $R(x)$ exhibits $2^d$ distinct modes in the $n$-dimensional HyperGrid, each located at a corner of the grid (represented by the keys). A two-dimensional illustration of the task with horizon $H$ is shown in Figure~\ref{fig:grid_example}, where the intensity of the color shading represents the magnitude of the rewards, with darker colors indicating higher reward values.
The objective for the agent is to model the distribution of the target rewards and successfully capture all the modes of the reward function. 

We evaluate the performance of each algorithm using five different seeds ($0$-$4$), and the results are presented as the mean performance across these runs, along with the corresponding standard deviation.
The implementation of all baseline methods is based on the publicly available open-source code\footnote{\url{https://github.com/GFNOrg/gflownet}} following default hyperparameters 
 as used in~\cite{bengio2021flow,malkin2022trajectory,madan2023learning}.
A detailed description of the reward function $R(x)$ and hyperparameter settings can be found in Appendix~\ref{app:gridWorld} due to space limitation.

\subsubsection{Performance Comparison}
We evaluate our method and compare it against strong baselines including Flow Matching (FM)~\citep{bengio2021flow}, Detailed Balance (DB)~\citep{bengio2023gflownet}, Trajectory Balance~\citep{malkin2022trajectory}, and Sub-Trajectory Balance~\citep{madan2023learning}, following the evaluation scheme in~\citep{bengio2021flow} based on the empirical $L_1$ error and the number of discovered modes.
Specifically, the $L_1$ error is computed as $\mathbb{E}[|p(x)-\pi(x)|]$, where $p(x)=R(x)/Z$ represents the underlying true reward distribution.
To estimate $\pi$, we repeatedly sample and summarize the visitation frequencies for each possible state $x$.
In this toy environment with horizon $H=16$ and increasing dimensions $d \in \{2, 3, 4\}$, the state space is relatively small, so the true reward distribution can be directly calculated since it allows for enumerating all possible states.
We measure the number of modes discovered during the last $1024$ samples.
The purpose of evaluating this toy environment is to verify that our method can converge to sampling proportionally to the rewards, and comparisons in more complex and practical environments are analyzed in Sections~\ref{sec:rna} and \ref{sec:mol}.

\begin{figure}[!h]
\centering
\subfloat[Small.]{\includegraphics[width=0.167\linewidth]{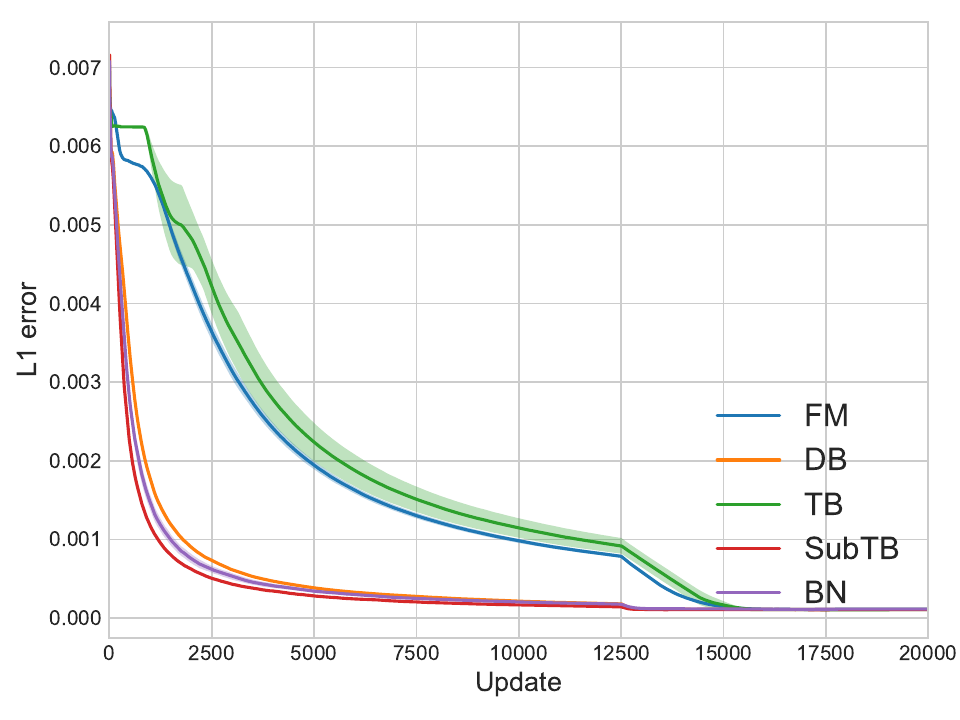}}
\subfloat[Medium.]{\includegraphics[width=0.167\linewidth]{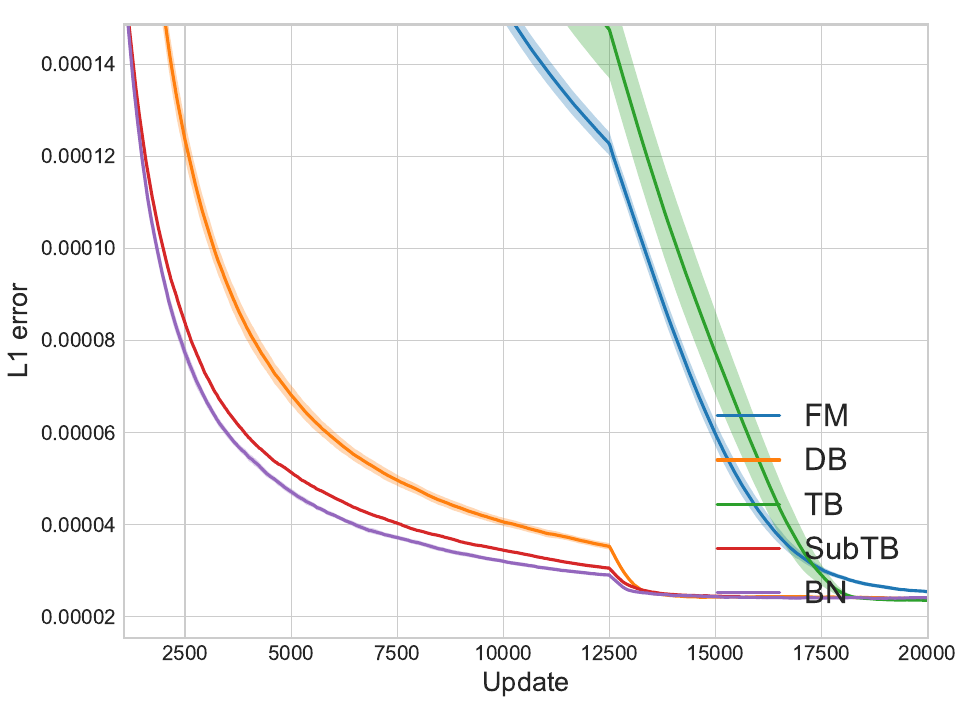}}
\subfloat[Large.]{\includegraphics[width=0.167\linewidth]{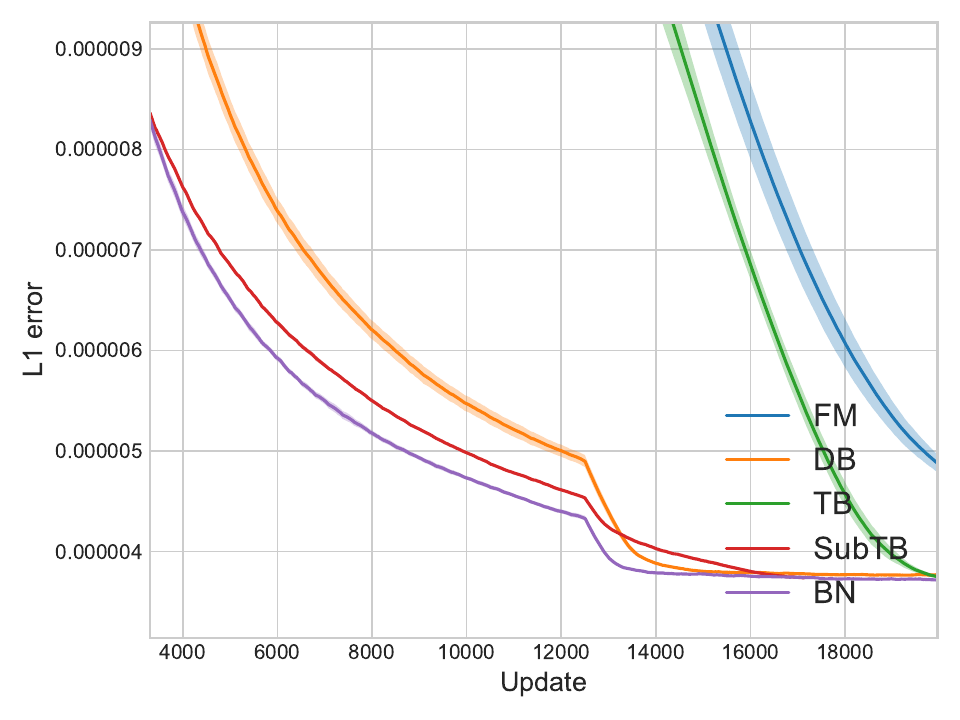}} 
\subfloat[Small.]{\includegraphics[width=0.167\linewidth]{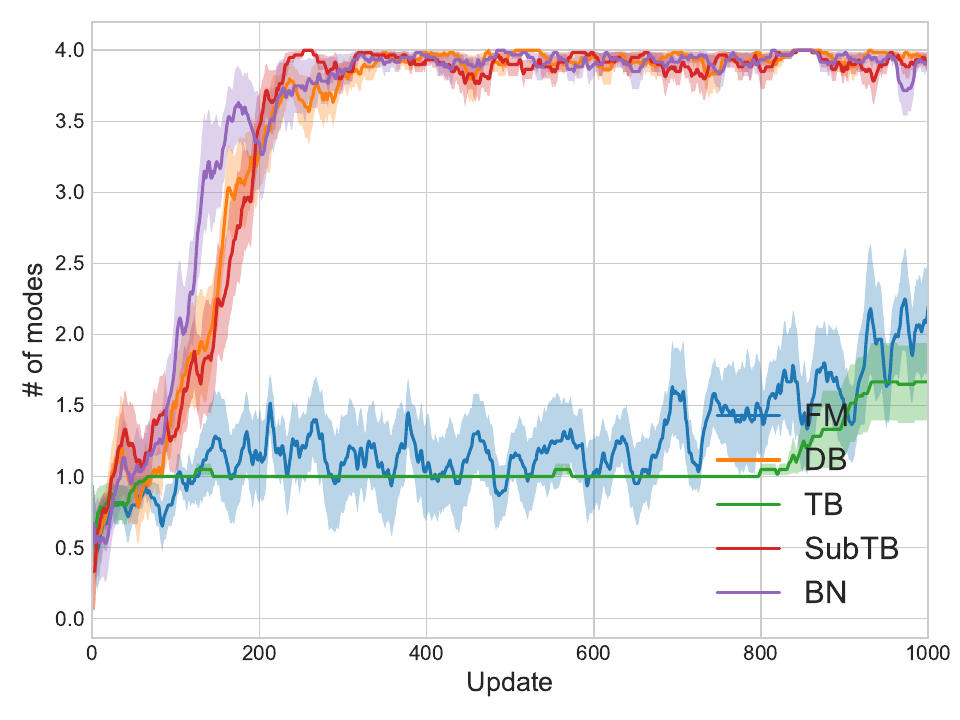}}
\subfloat[Medium.]{\includegraphics[width=0.167\linewidth]{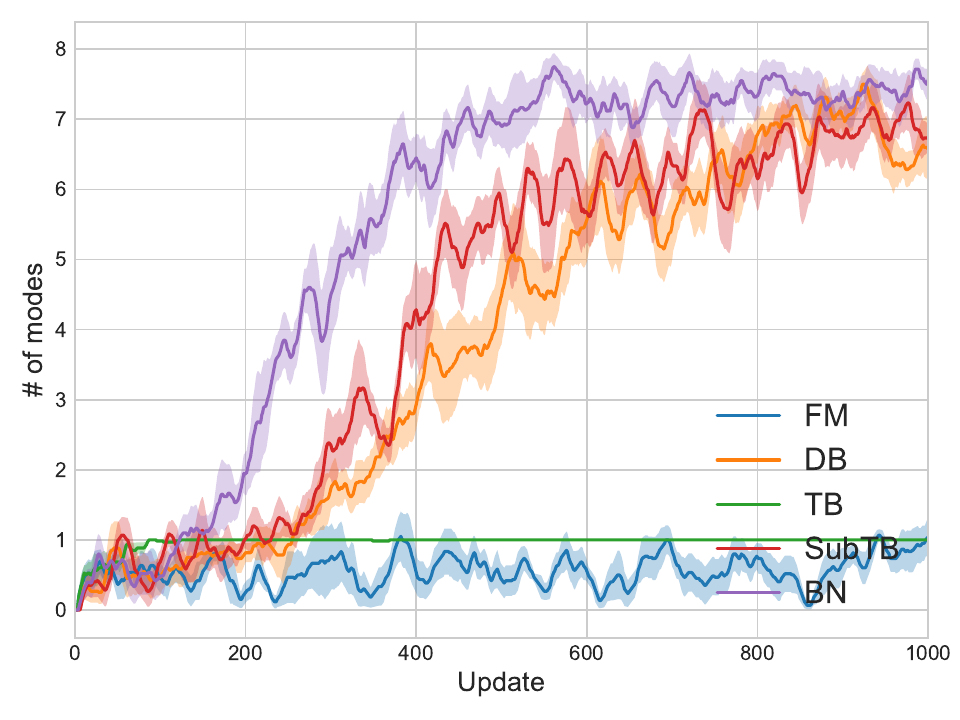}}
\subfloat[Large.]{\includegraphics[width=0.167\linewidth]{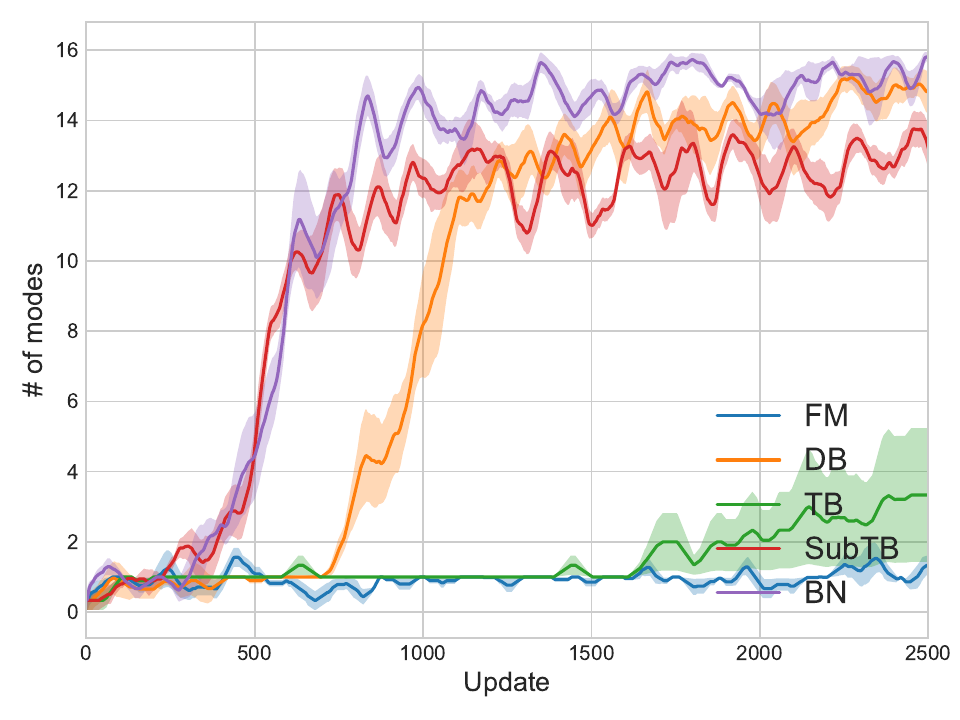}}
\caption{Performance comparison on HyperGrid in terms of number of updates. (a)-(c): Empirical $L_1$ error. (d)-(f): Number of modes.}
\label{fig:grid}
\end{figure}

\begin{figure}[!h]
\centering
\subfloat[Small.]{\includegraphics[width=0.167\linewidth]{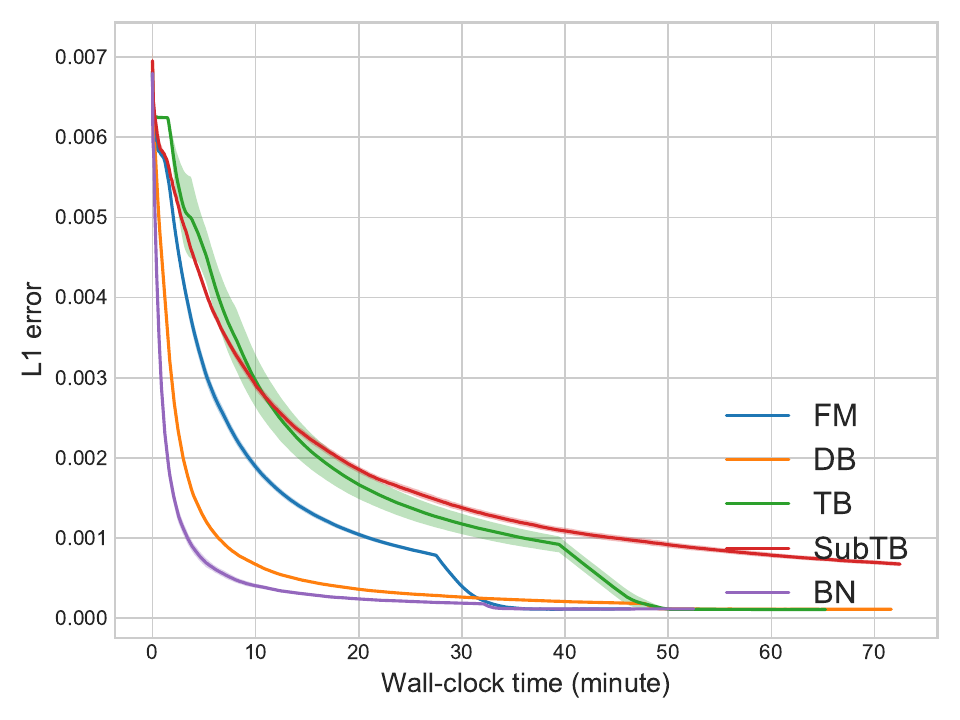}}
\subfloat[Medium.]{\includegraphics[width=0.167\linewidth]{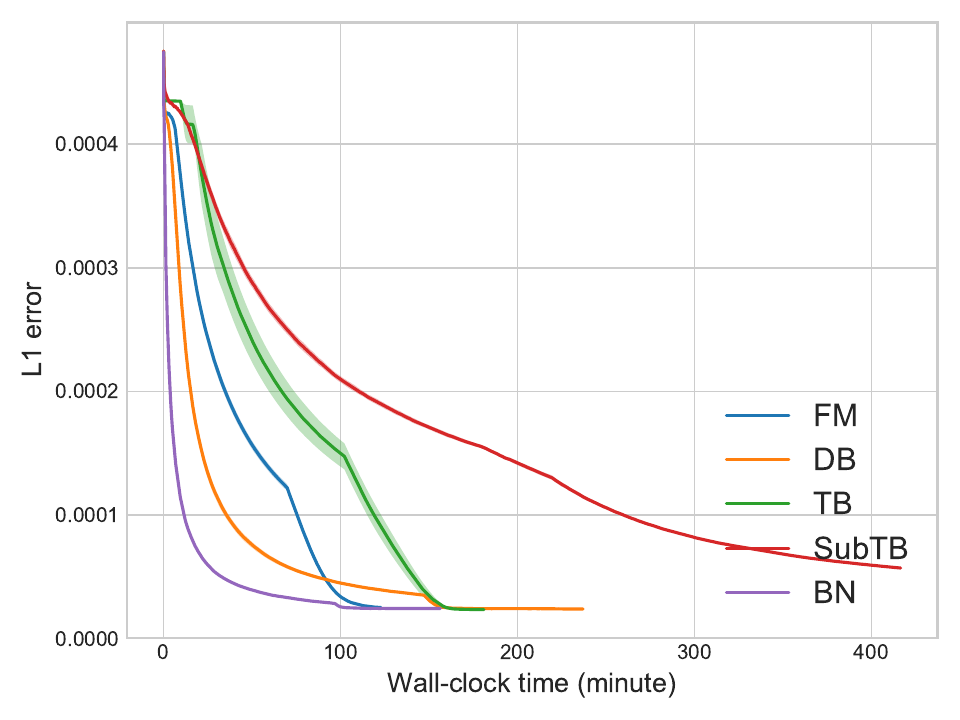}}
\subfloat[Large.]{\includegraphics[width=0.167\linewidth]{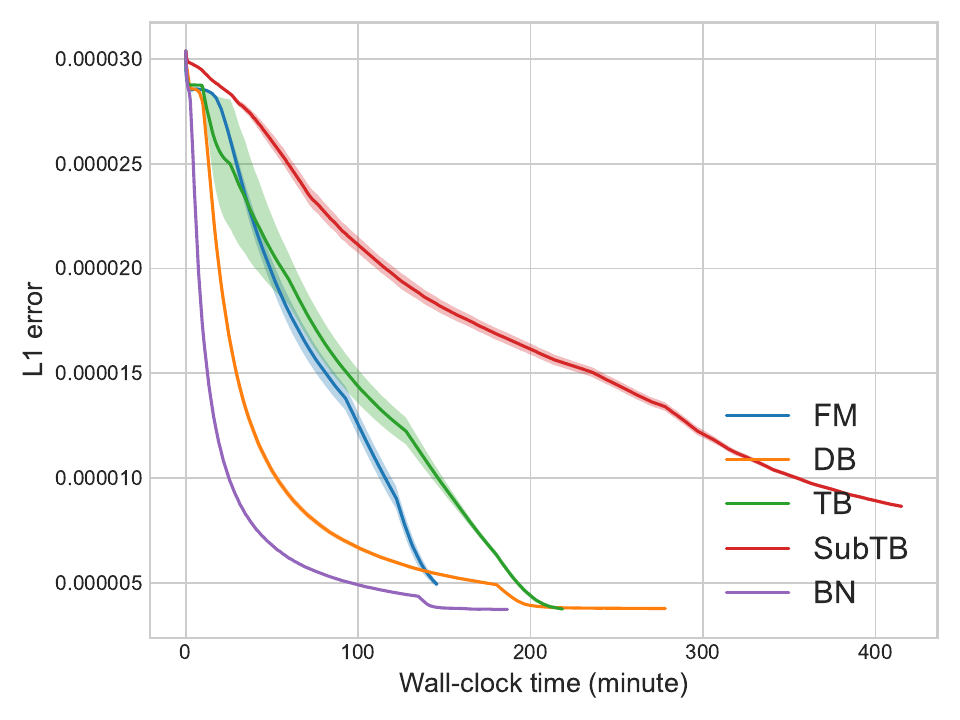}} 
\subfloat[Small.]{\includegraphics[width=0.167\linewidth]{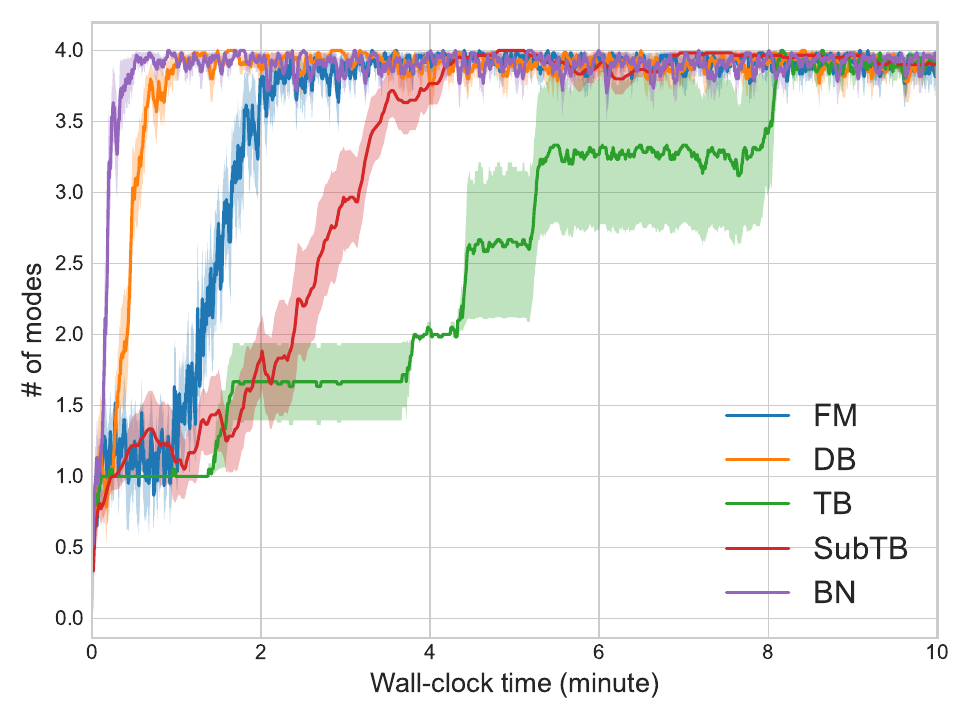}}
\subfloat[Medium.]{\includegraphics[width=0.167\linewidth]{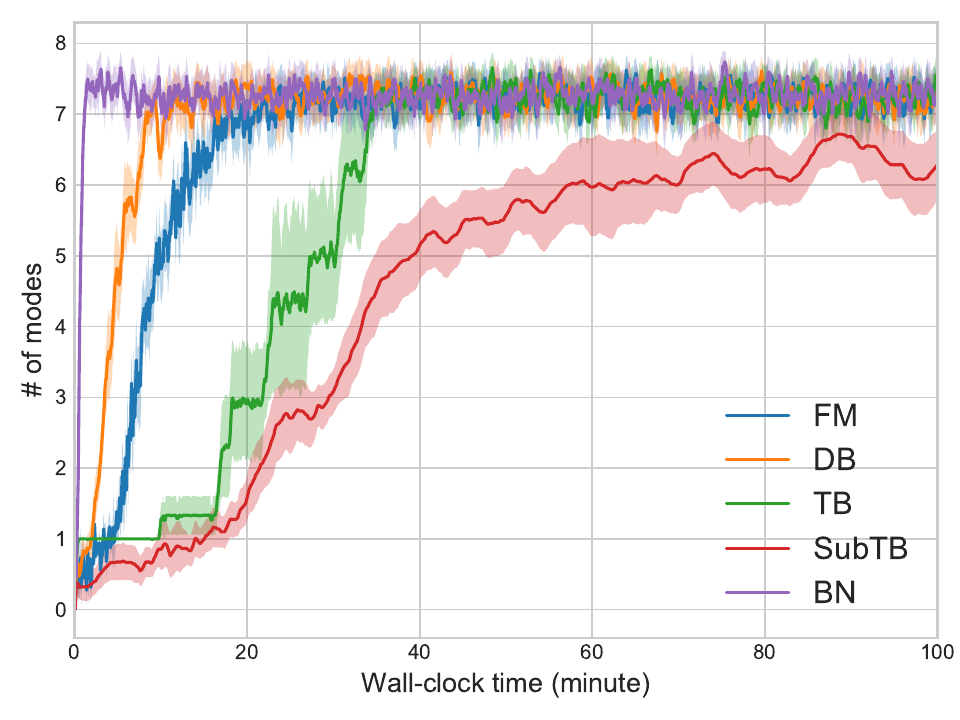}}
\subfloat[Large.]{\includegraphics[width=0.167\linewidth]{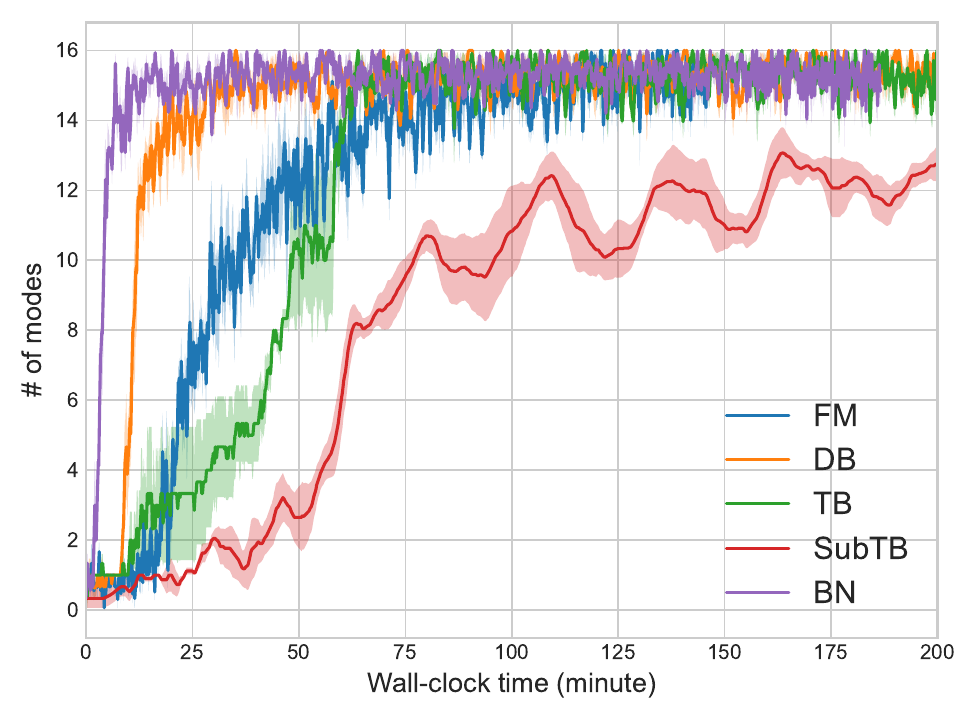}}
\caption{Performance comparison on HyperGrid in terms of wall-clock time. (a)-(c): Empirical $L_1$ error. (d)-(f): Number of modes.}
\label{fig:grid_time}
\end{figure}

The comparison results in HyperGrid in terms of the number of updates~\citep{bengio2021flow} with increasing dimensions of the action space $n$ (from small, medium to large) are summarized in Figures~\ref{fig:grid}(b)-(d). 
As shown, our approach significantly improves the learning efficiency of FM, which validates our hypothesis that the well-designed learning structure improves data efficiency. Furthermore, our method outperforms strong baselines including DB and TB by a large margin.

In this toy environment, the most competitive baseline with our approach is SubTB, where our BN method performs comparably to this strong baseline and outperforms it as the problem scale increases with larger action spaces. However, it is important to note that the computational cost of SubTB grows quadratically with the trajectory length $n$, as it considers all possible O($n^2$) sub-trajectories when computing the loss for a single trajectory. In contrast, our method only grows linearly with the trajectory length.
We further compare each method in terms of wall-clock time as studied in~\citep{falet2023delta}.
As shown in Figure~\ref{fig:grid_time}, BN significantly outperforms SubTB, which is more computationally efficient and leads to faster wall-clock time convergence.

\subsection{Biological Sequence Design} \label{sec:rna}
\subsubsection{Experimental Setup}
We now study a practical task of generating RNA sequences consisting of $14$ nucleobases which aims to achieve high binding affinity to the target transcription factor.
At each timestep, the agent chooses to either prepend or append a token to the current state, following the setup in~\citep{shen2023towards} that results in a directed acyclic graph instead of a simple tree. 
We consider four different target transcriptions introduced by~\citep{lorenz2011viennarna}.
Further details for the experimental setup can be found in Appendix~\ref{app:RNA_seq_generation}.

\subsubsection{Results}
We follow the evaluation scheme in~\citet{shen2023towards,kim2023local} and evaluate our method and baselines in terms of accuracy, which measures how well it matches the target reward distribution. Specifically, it is computed by using a relative error between the sample mean of $R(x)$ under the learned policy distribution $P_F(x)$ and the expectation of $R(x)$ given the target distribution $R(x)/Z$, i.e., $\frac{\mathbb{E}_{P_F(x)}[R(x)]}{\mathbb{E}_{p*(x)}[R(x)]}$ (and clipped with a maximum value of $1$).
In addition, we also measure the number of modes discovered by each method during the course of learning following~\citet{shen2023towards,kim2023local}, for investigate the diversity-seeking ability of each method.

\begin{figure}[!h]
\centering
\subfloat[L14\_RNA1.]{\includegraphics[width=0.25\linewidth]{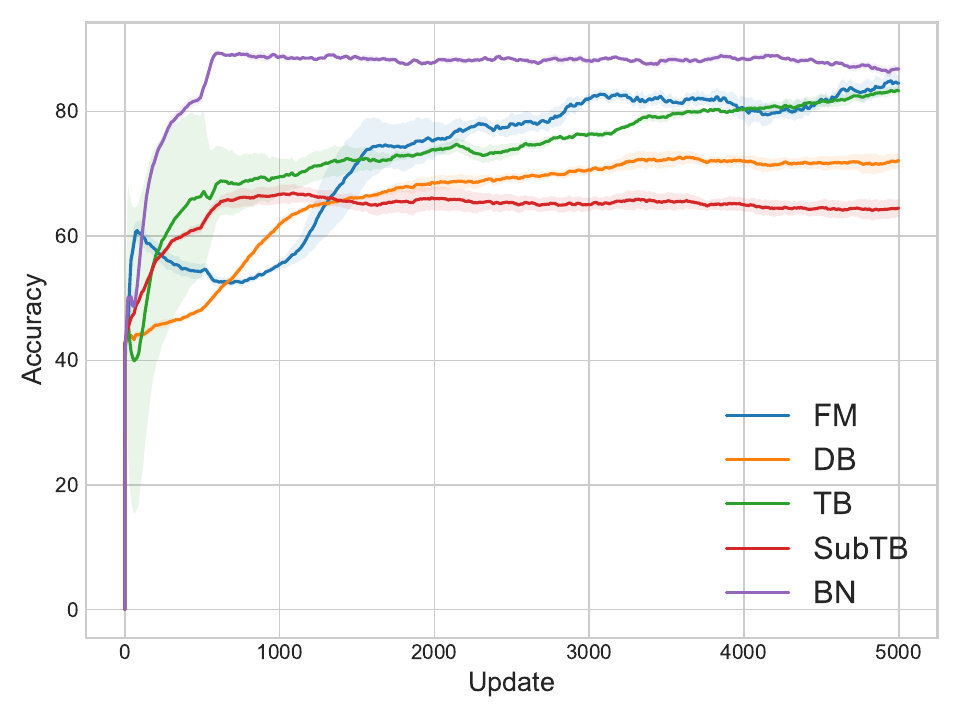}}
\subfloat[L14\_RNA2.]{\includegraphics[width=0.25\linewidth]{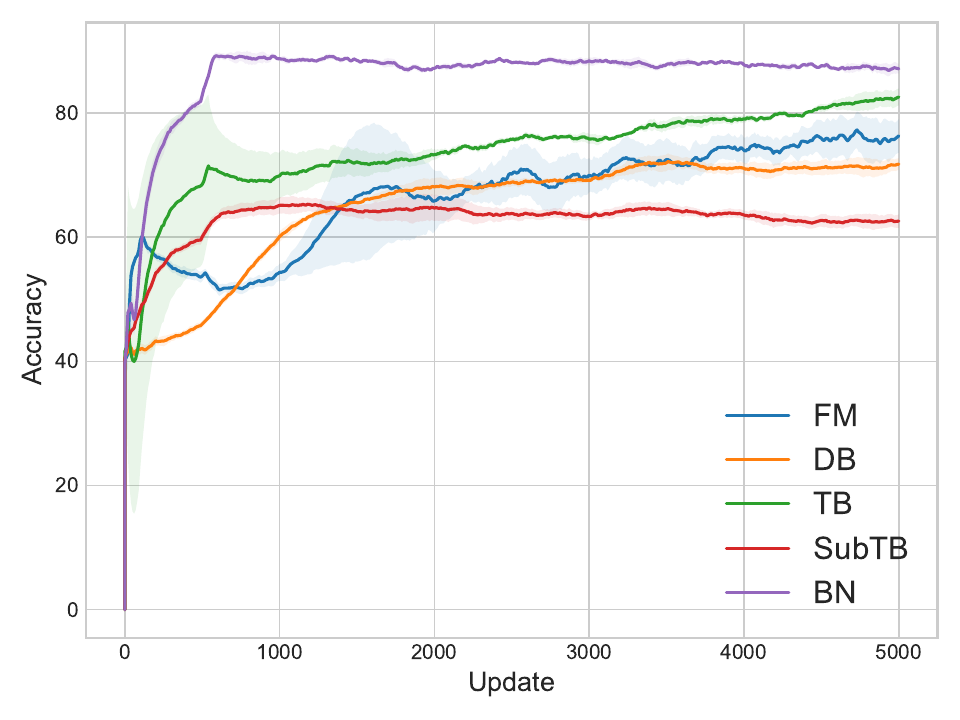}}
\subfloat[L14\_RNA3.]{\includegraphics[width=0.25\linewidth]{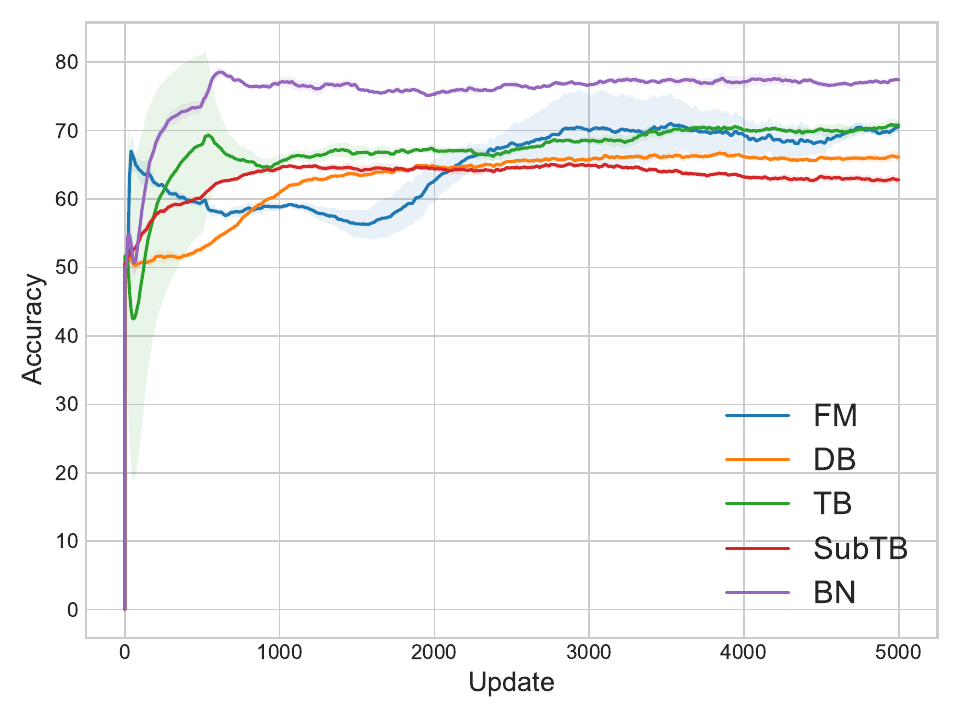}}
\subfloat[L14\_RNA4.]{\includegraphics[width=0.25\linewidth]{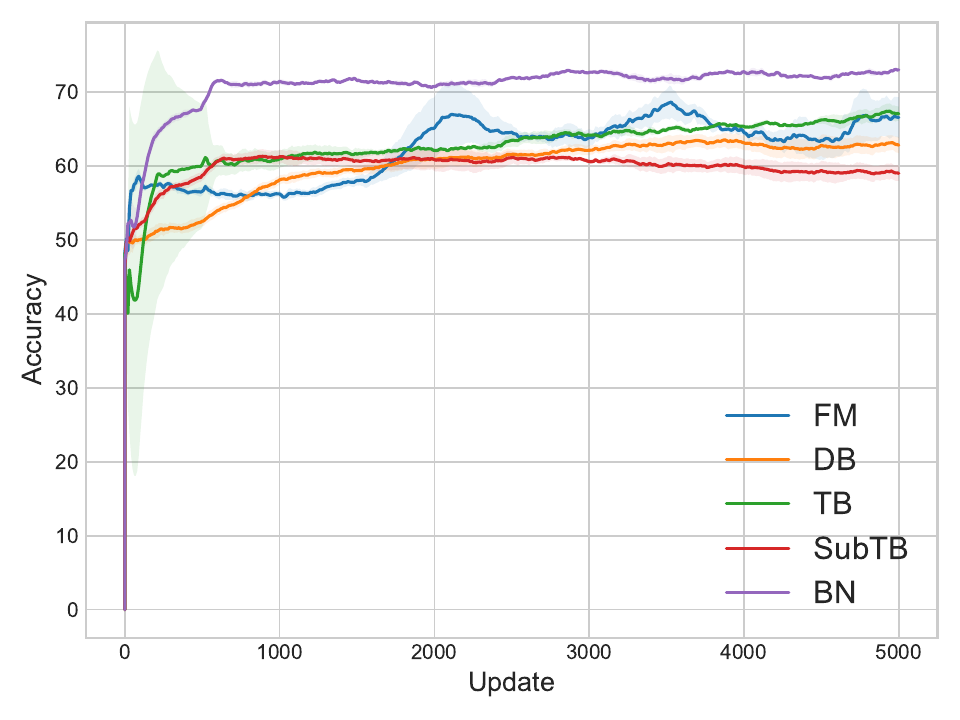}}
\caption{Comparison in terms of accuracy for different methods in RNA generation tasks.}
\label{fig:rna_acc}
\end{figure}

\begin{figure}[!h]
\centering
\subfloat[L14\_RNA1.]{\includegraphics[width=0.25\linewidth]{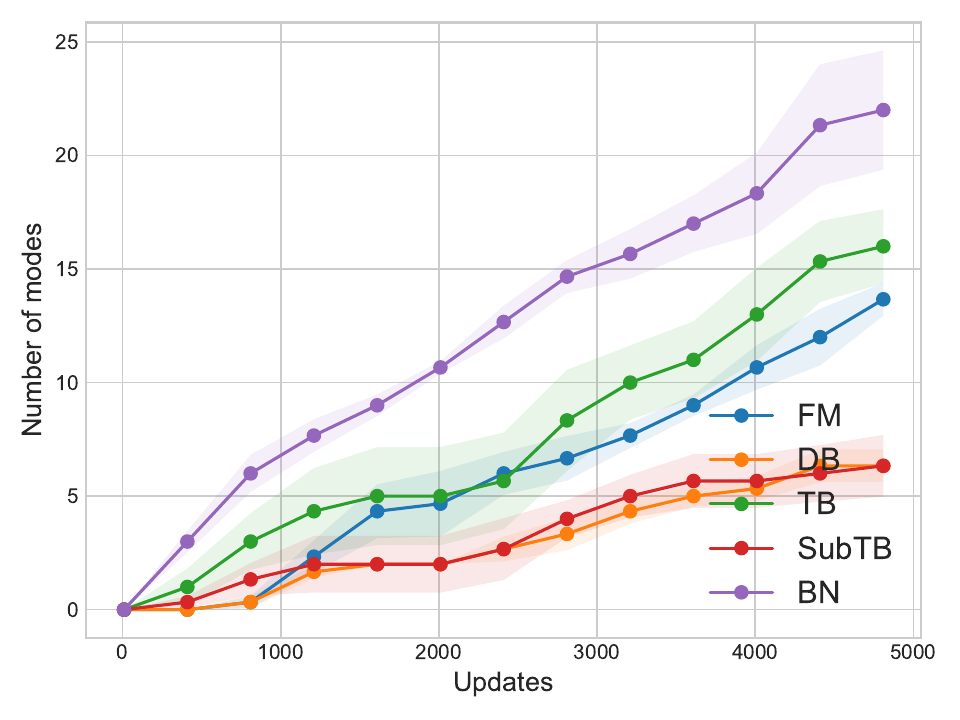}}
\subfloat[L14\_RNA2.]{\includegraphics[width=0.25\linewidth]{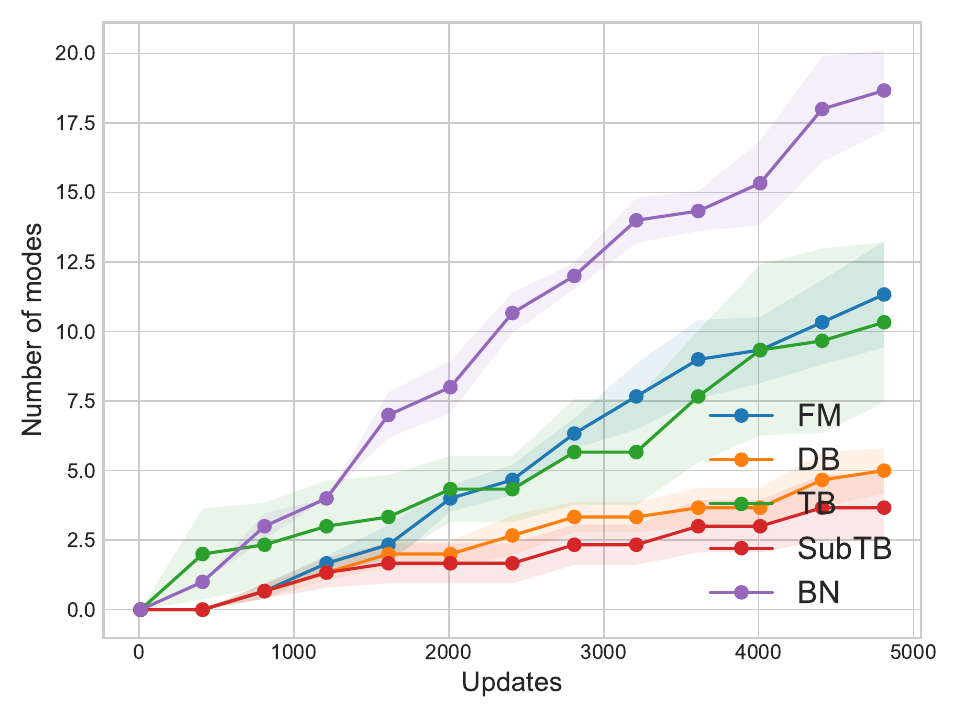}}
\subfloat[L14\_RNA3.]{\includegraphics[width=0.25\linewidth]{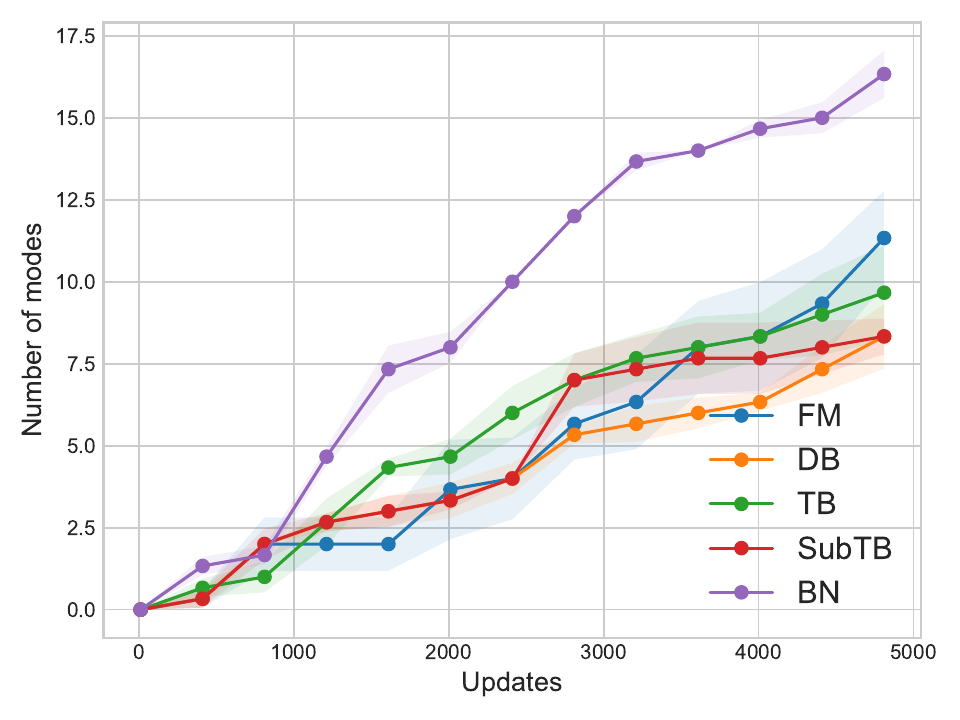}}
\subfloat[L14\_RNA4.]{\includegraphics[width=0.25\linewidth]{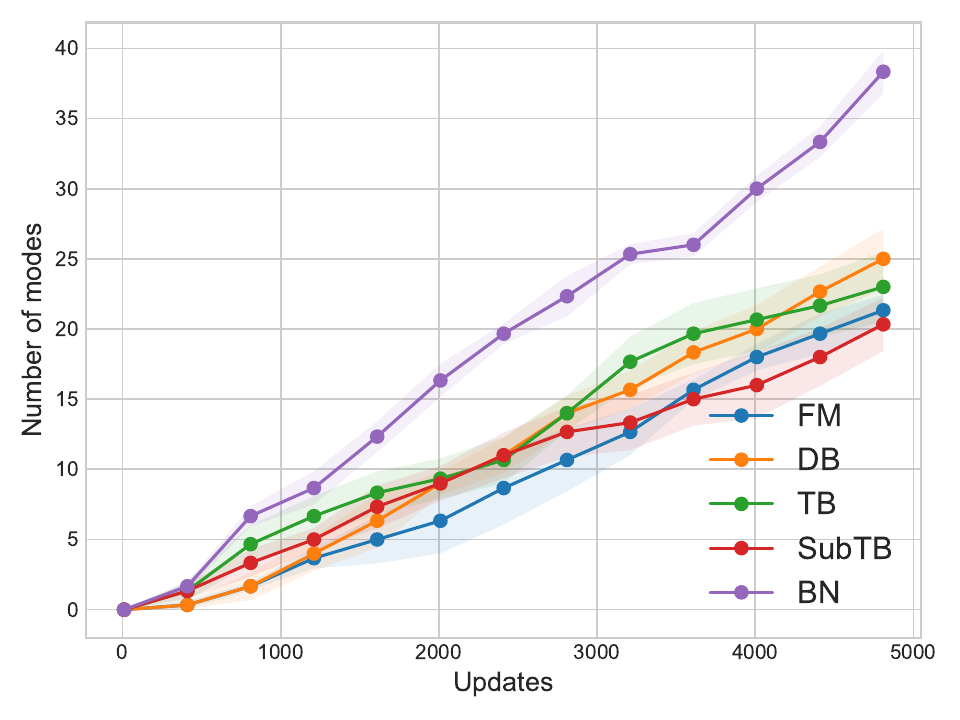}}
\caption{Comparison in terms of the number of modes discovered for different methods in RNA generation tasks.}
\label{fig:rna_mode}
\end{figure}

As demonstrated in Figure~\ref{fig:rna_acc}, our method achieves higher accuracy and learns more efficiently than other baselines in different reward setups.
Figure~\ref{fig:rna_mode} demonstrates the number of modes discovered by each method, which shows that our proposed approach discovers more modes in a more efficiently manner, which validates its effectiveness in practical tasks.

\subsection{Molecule Generation} \label{sec:mol}
\subsubsection{Experimental Setup}
In this section, we investigate the efficacy of our approach on the more complex and larger-scale molecule generation task~\citep{bengio2021flow}, 
In this task, molecules are represented as graphs, composed of a set of building blocks. 
The agent iteratively constructs the molecule by deciding the attachment point and the specific block to attach at each step, while adhering to chemical validity constraints. 
The action space also includes an exit action allowing the agent to terminate the generation process. 
This task presents significant challenges due to the large state (approximately $10^6$) and action (can contain about $2000$ actions) spaces. 
The objective of the agent is to discover a diverse set of molecules with high rewards, corresponding to low binding energy to the soluble epoxide hydrolase (sEH) protein~\citep{bengio2021flow}. 
Following~\citet{bengio2021flow}, we employ a pre-trained proxy model to estimate this binding energy. 
Further details regarding the experimental setup can be found in Appendix~\ref{app:molecule_generation}.

\subsubsection{Results}
\begin{figure}[!h]
\centering
\subfloat[]{\includegraphics[width=0.25\linewidth]{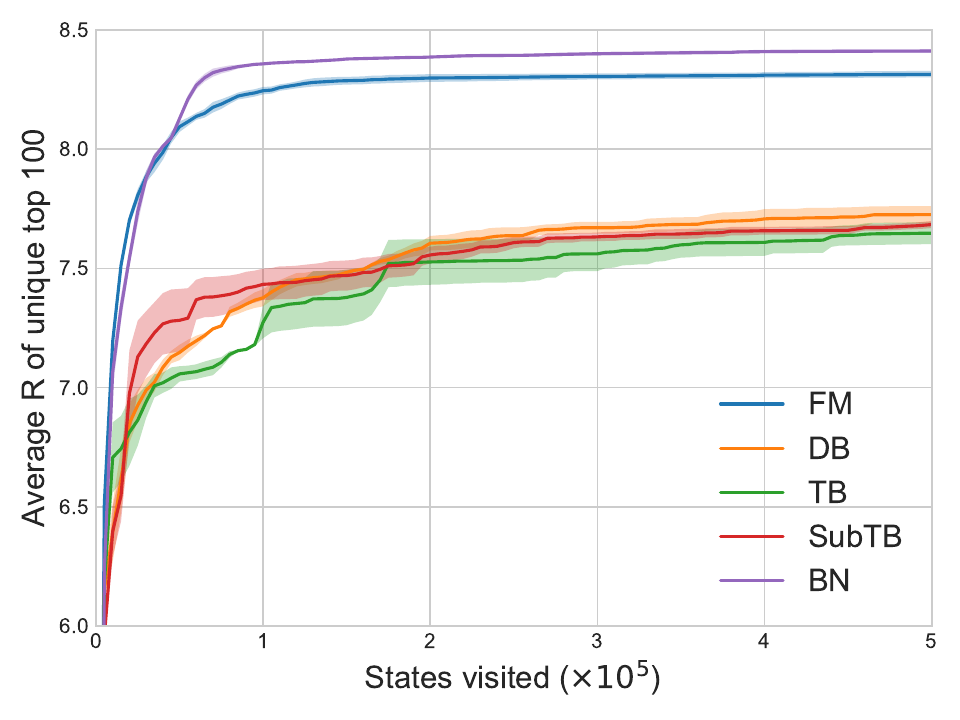}}
\subfloat[]{\includegraphics[width=0.25\linewidth]{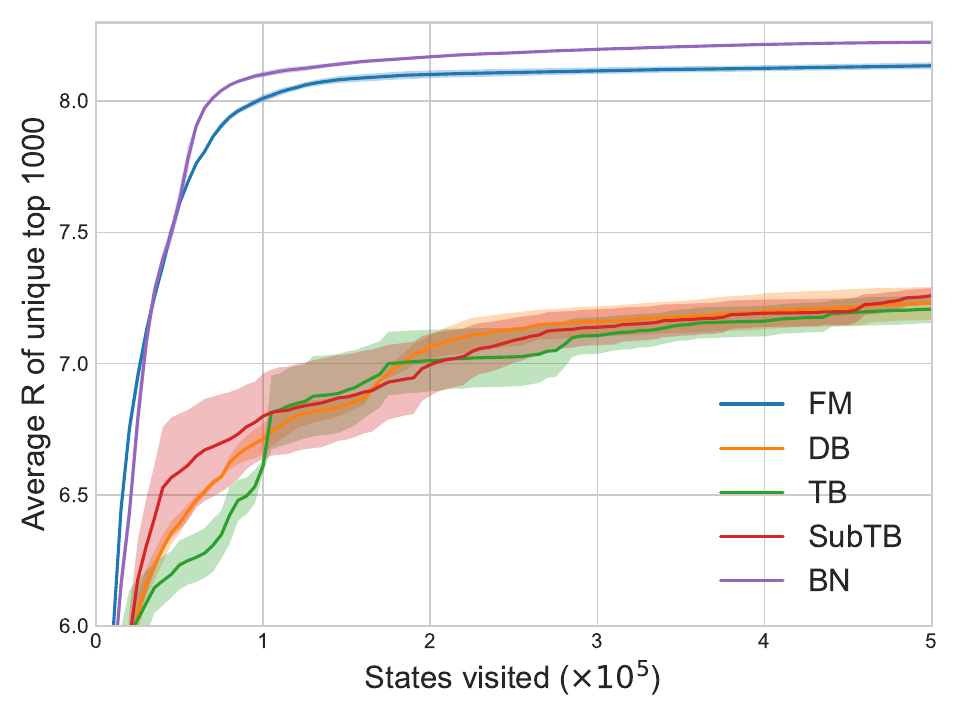}}
\subfloat[]{\includegraphics[width=0.25\linewidth]{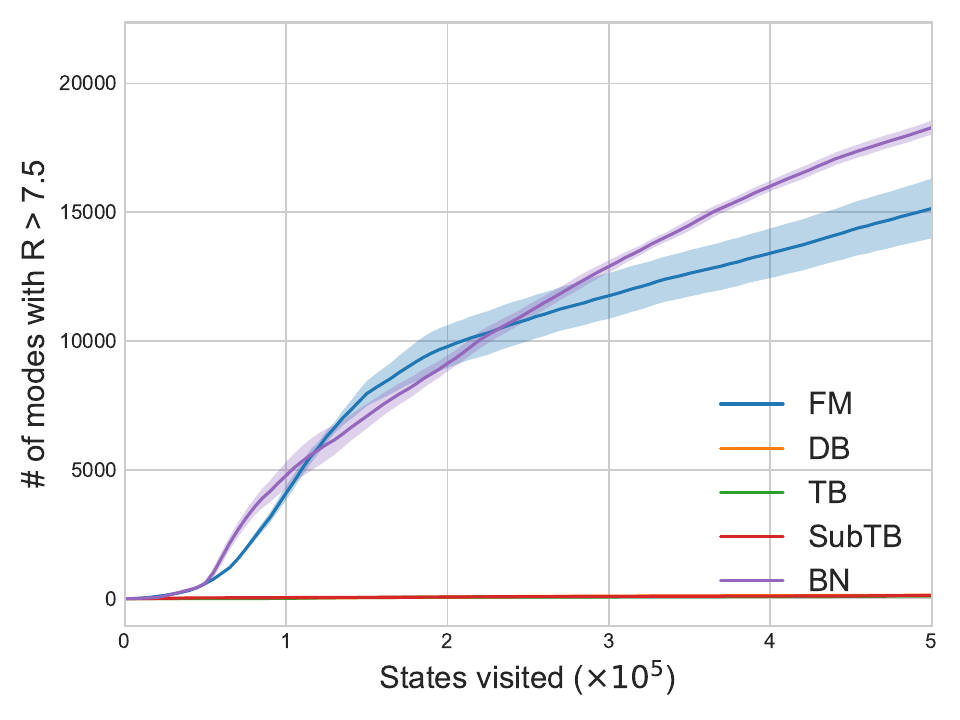}}
\subfloat[]{\includegraphics[width=0.25\linewidth]{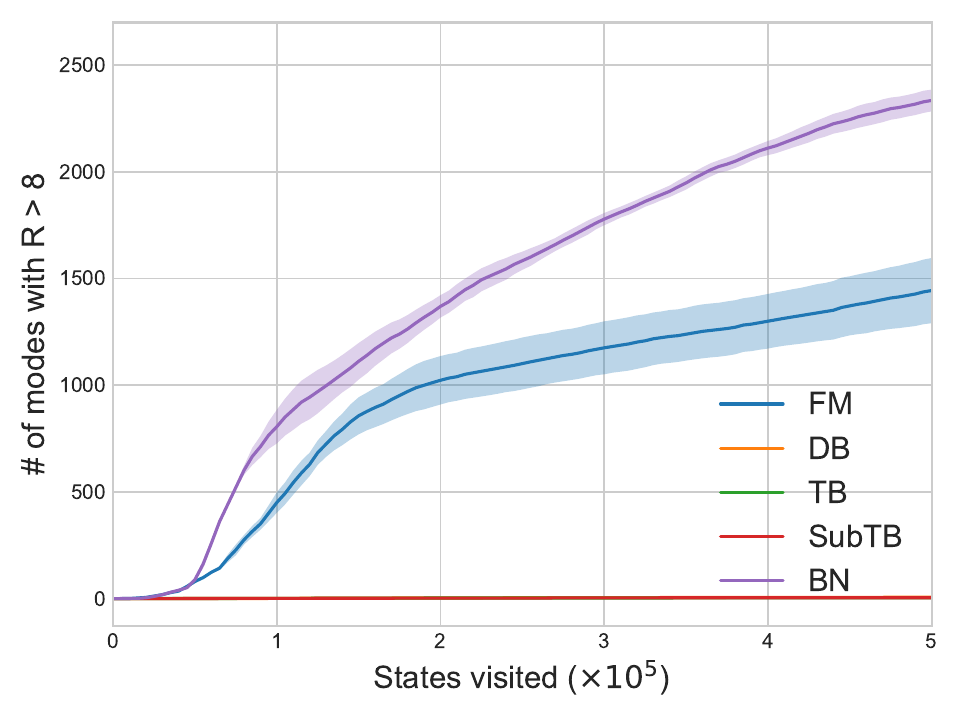}}
\caption{Results in molecule generation. (a) Top-$100$ rewards. (b) Top-$1000$ rewards. (c) Number of modes with $R > 7.5$. (d) Number of modes with $R > 8.0$.}
\label{fig:mol_res}
\end{figure}

We follow the same evaluation scheme as in~\citet{bengio2021flow} by comparing each method in terms of top-$K$ rewards and the number of modes discovered during the training course.
As shown in Figure~\ref{fig:mol_res}(a)-(b), our method achieves the highest top-$K$ rewards, where the most competitive baseline in this large-scale problem is FM (consistent to previous studies~\citep{pan2022generative,zhang2023distributional}). Other baselines that involve learning or specifying a backward policy fail to perform well. 
In addition, our method also discovers many more modes than FM, with a significant gap compared to DB, TB, and SubTB. 
Table~\ref{tab:mol} summarizes the number of new unique molecules discovered by each method with a score above $8$ that
are not contained in the training dataset.
As demonstrated, our method has the potential to discover new candidates with high quality that have not been seen before.
We further compute the average pairwise Tanimoto similarity~\citep{bengio2021flow} for the top-$K$ molecules generated by BN and the most competitive FM method, where BN has a result of $0.472 \pm 0.015$ and FM has a result of $0.497 \pm 0.037$, which indicates that the high-performing samples discovered by our proposed method are more diverse than FM.
The results demonstrate the effectiveness of BN in large-scale practical problems.

\begin{table}[!h]
\centering
\caption{The number of new unique molecules discovered by each method with a score above $8$ that are not contained in the training dataset.}
\label{tab:mol}
\begin{tabular}{cccccc}
\toprule 
Method & DB & TB & SubTB & FM & BN \\ 
\midrule 
$\#$ of mols & $6.67 \pm 4.11$ & $4.33 \pm 2.05$ & $4.67 \pm 2.62$ & $1447.00 \pm 267.55$ & $\mathbf{2345.67 \pm 92.07}$ \\
\bottomrule    
\end{tabular}
\end{table}

\section{Conclusion}

In this paper, we introduce Bifurcated GFlowNets (BN), a novel approach to address the challenges of data efficiency and scaling up to larger-scale problems. BN factorizes edge flows into separate representations for state flows and edge-based allocations, which leads to more efficient learning. Extensive experiments on standard benchmarks demonstrated that BN outperforms existing GFlowNet variants, achieving superior performance in terms of both learning efficiency and overall effectiveness. It will be interesting for future work to apply our BN approach to other applications such as combinatorial optimization.
It will be interesting for future work to apply our approach to combinatorial optimization, large language models, and other complex systems with huge action spaces.

\bibliography{main}
\bibliographystyle{plainnat}

\clearpage
\appendix

\section{Proofs}
\label{app:proof}
\textbf{Theorem 4.1}
\emph{If $\mathcal{L}_{\text{BN}}(s')=0$ for all states, then the edge advantage policy $A(s'|s)$ samples proportionally to the reward function.}

\begin{proof}
As for all state $s'$, $\mathcal{L}_{\text{BN}}(s')=0$, we have that
\begin{equation}
\forall s', \sum_{s \to s' \in \mathcal{A}}F(s)A(s'|s)=F(s').
\end{equation}

Let $F(s \to s')=F(s)A(s'|s)$, then we have that
\begin{equation}
\label{left}
\forall s',\sum_{s \to s' \in \mathcal{A}}F(s \to s')=\sum_{s \to s' \in \mathcal{A}}F(s)A(s'|s)=F(s').
\end{equation}
We also have that
\begin{equation}
\forall s', \sum_{s' \to s'' \in \mathcal{A}}F(s' \to s'')=\sum_{s' \to s'' \in \mathcal{A}}F(s')A(s''|s').
\end{equation}

As $\forall s', \sum_{s' \to s''\in \mathcal{A}}A(s''|s')=1$, 
we have that
\begin{equation}
\label{right}
\forall s', \sum_{s' \to s'' \in \mathcal{A}}F(s' \to s'')=F(s').
\end{equation}

Combing eq.(\ref{left}) with eq.(\ref{right}), we get that
\begin{equation}
\forall s', \sum_{s \to s' \in \mathcal{A}}F(s \to s')=F(s')=\sum_{s' \to s'' \in \mathcal{A}}F(s' \to s'').
\end{equation}

Then, the edge flow function $F(s \to s')$ and the state flow function $F(s')$ satisfy the flow matching condition. As $F(s \to s')=F(s)A(s'|s)$, we get the sampling policy $A(s'|s)=F(s \to s')/F(s)$ does samples the objects proportionally to the reward function.
\end{proof}

\section{Experimental Setup}
\label{app:exp set up}

All baseline methods are implemented based on the open-source implementation\footnote{\url{https://github.com/GFNOrg/gflownet}}\footnote{\url{https://github.com/dbsxodud-11/ls_gfn}}, following the default hyper-parameters and setup as described in~\citep{bengio2021flow}. For computational resources, all experiments are conducted on the NVIDIA RTX 3090.

\subsection{ToyDAG}
\label{app:setup_toyDAG}

\begin{figure}[!h]
\centering
\subfloat[Small DAG]{\includegraphics[width=0.35\linewidth]{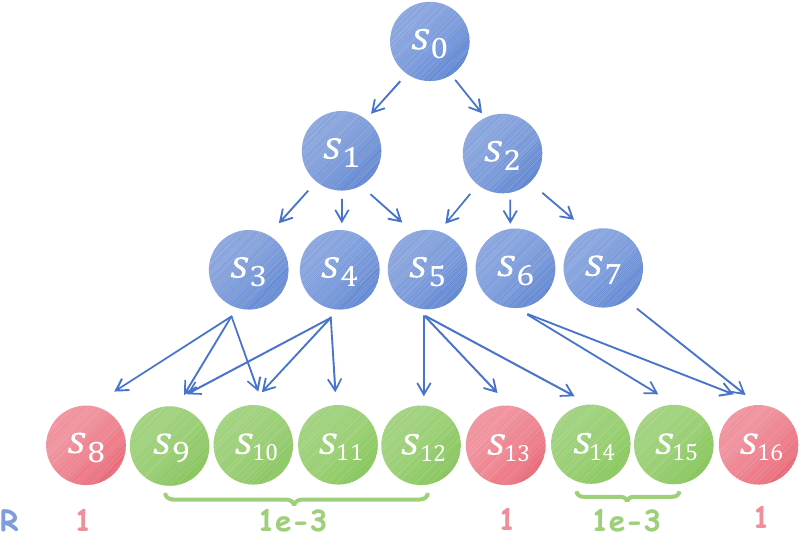}}
\subfloat[Large DAG]{\includegraphics[width=0.35\linewidth]{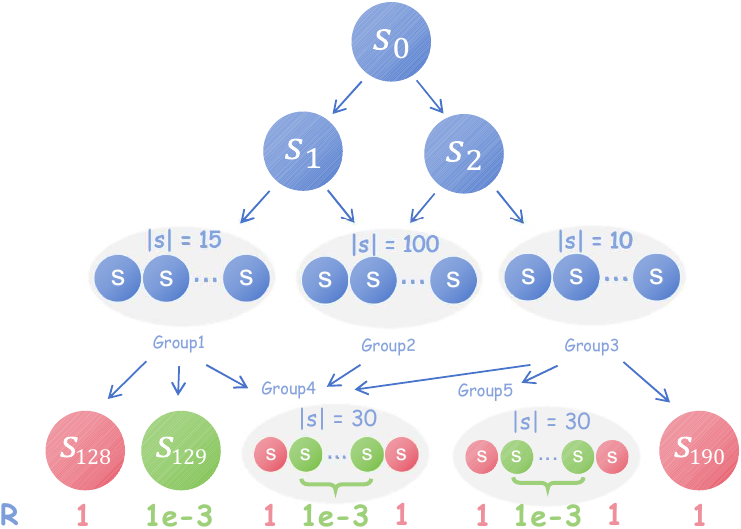}}
\caption{Detailed illustration of the DAG in section~\ref{sec:empirical validation}.}
\label{fig:explain DAG}
\end{figure}

Figure~\ref{fig:explain DAG}(a)-(b) illustrate the didactic environment designed in Section~\ref{sec:empirical validation}, representing the small and large versions, respectively. Figure~\ref{fig:explain DAG}(a) depicts the small version, where both the state and action space sizes are small. Here, states are denoted by $s$, blue arrows represent state transitions, blue circles represent intermediate states, and the last row shows terminal states. Red circles indicate a reward of $R = 1$, while green circles indicate a reward of $R = 1e-3$.

Figure~\ref{fig:explain DAG}(b) illustrates the large version, which has larger state and action spaces. Different state groups are represented by $Group$, with each group containing multiple states and we mark the number of states included. Arrows pointing into or out of a state group apply to every state within that group. For example, the transition $s_1 \rightarrow Group_1$ indicates that state $s_1$ has $size(Group_1) = 15$ actions, each leading to a different state within $Group 1$. The transition $Group_1 \rightarrow s_{128}$ indicates that every state within $Group 1$ has one action pointing to $s_{128}$. Similarly, the transition $Group_1 \rightarrow Group_5$ means that each state within $Group_1$ has $size(Group_5) = 30$ actions, each leading to a different state within $Group_5$. The last row illustrates the terminal states and their corresponding rewards, with different rewards distinguished by color. Notably, in $Group_4$ and $Group_5$, the reward of the first and last states is $1$, while the remaining states are $1e-3$. 

\subsection{GridWorld}
\label{app:gridWorld}
For the GridWorld task, we employed the standard reward function described by~\citep{bengio2021flow} $R(x) = \frac{1}{2} \prod_{i} \mathbb{I} (0.25 < |x_i/H - 0.5|) + 2 \prod_{i} \mathbb{I} (0.3 < |x_i/H - 0.5| < 0.4) + 10^{-6}$.
This reward distribution, illustrated in Figure~\ref{fig:grid_example}, features four modes located near the corners of the maze. Our GFlowNets architecture for this task comprises two hidden layers, with 256 hidden units per layer using Leaky-ReLU activation. 
In the training process, we use the Adam optimizer~\citep{kingma2014adam} with a learning rate 0.001 and a batch size set to 16 for training iterations of 20,000.

\subsection{RNA Sequence Generation}
\label{app:RNA_seq_generation}
In the RNA Sequence Generation task, the objective is to generate RNA sequences of 14 nucleobases with strong binding affinity to a specific transcription factor. At each step, the agent appends or prepends a nucleobase following the approach outlined by~\citep{lorenz2011viennarna}, constructing a Directed Acyclic Graph (DAG). The RNA sequence generation process relies on a nucleobase vocabulary of four options. Our GFlowNets for this task comprise two hidden layers, each containing 2048 units and employing ReLU activation functions. The exploration strategy adopts an $\epsilon$-greedy approach with $\epsilon$ set to 0.001, while the reward exponent remains fixed at 3. Training is conducted with a learning rate of 1e-4 and a batch size of 32 for 5000 iterations.

\subsection{Molecule Generation}
\label{app:molecule_generation}
For the Molecule Generation task, rewards are evaluated using a pre-trained proxy model trained on a dataset of 300,000 randomly generated molecules as described in~\citep{bengio2023gflownet}. Rewards are determined based on normalized scores, where the agent receives non-zero rewards only if the normalized score exceeds a specified threshold of 7.0; Otherwise, the reward remains zero. The agent's selections are drawn from a vocabulary of 105 fundamental building blocks to extend the molecule. Training runs for 80000 iterations, with a learning rate of 5e-4 and a batch size of 4.

\end{document}

%% file: math_commands.tex
\usepackage{amsmath,amsfonts,bm}

\def\eqref#1{equation~\ref{#1}}

\def\1{\bm{1}}

\DeclareMathAlphabet{\mathsfit}{\encodingdefault}{\sfdefault}{m}{sl}
\SetMathAlphabet{\mathsfit}{bold}{\encodingdefault}{\sfdefault}{bx}{n}

